\documentclass[journal,transmag]{IEEEtran}
\usepackage{caption}
\usepackage{amsmath}
\usepackage{amssymb}
\usepackage{booktabs}
\usepackage{array}
\usepackage[final]{graphicx}
\usepackage{float}
\usepackage[caption = false]{subfig}
\ifCLASSINFOpdf
\else
\fi
\begin{document}
\title{Multi-Modal Attention-based Fusion Model \\ for Semantic Segmentation of RGB-Depth Images} 


\author{\IEEEauthorblockN{Fahimeh Fooladgar, and
Shohreh Kasaei,~\IEEEmembership{Senior Member,~IEEE}}
\IEEEauthorblockA{Department of Computer Engineering,
Sharif University of Technology,Tehran, Iran}}

%



%
\IEEEtitleabstractindextext{%
\begin{abstract}
  The 3D scene understanding is mainly considered as a crucial requirement in computer vision and robotics applications. One of the high-level tasks in 3D scene understanding is semantic segmentation of RGB-Depth images. With the availability of RGB-D cameras, it is desired to improve the accuracy of the scene understanding process by exploiting the depth features along with the appearance features. As depth images are independent of illumination, they can improve the quality of semantic labeling alongside RGB images. Consideration of both common and specific features of these two modalities improves the performance of semantic segmentation. One of the main problems in RGB-Depth semantic segmentation is how to fuse or combine these two modalities to achieve more advantages of each modality while being computationally efficient. Recently, the methods that encounter deep convolutional neural networks have reached the state-of-the-art results by early, late, and middle fusion strategies. In this paper, an efficient encoder-decoder model with the attention-based fusion block is proposed to integrate mutual influences between feature maps of these two modalities. This block explicitly extracts the interdependences among concatenated feature maps of these modalities to exploit more powerful feature maps from RGB-Depth images.  The extensive experimental results on three main challenging datasets of NYU-V2, SUN RGB-D, and Stanford 2D-3D-Semantic show that the proposed network outperforms the state-of-the-art models with respect to computational cost as well as model size. Experimental results also illustrate the effectiveness of the proposed lightweight attention-based fusion model in terms of accuracy.
  
\end{abstract}

}

\maketitle

\IEEEdisplaynontitleabstractindextext

%
\IEEEpeerreviewmaketitle

\section{Introduction}

\IEEEPARstart{S}{emantic} segmentation of RGB-Depth images has been considered as one of the main tasks for 3D scene understanding. The popularity of its applications such as autonomous driving, augmented virtual reality, and the inference of support relations among objects in robotics emphasize the importance of scene understanding. Most of the researches in this field have been done on outdoor scenes which are less challenging compared to indoor scenes. The existence of small objects, light-tailed distribution of objects, occlusions, and poor illumination cause major challenges in indoor scenes, to name a few.

\begin{figure*}
	\begin{center}
		\includegraphics[width=0.8\linewidth]{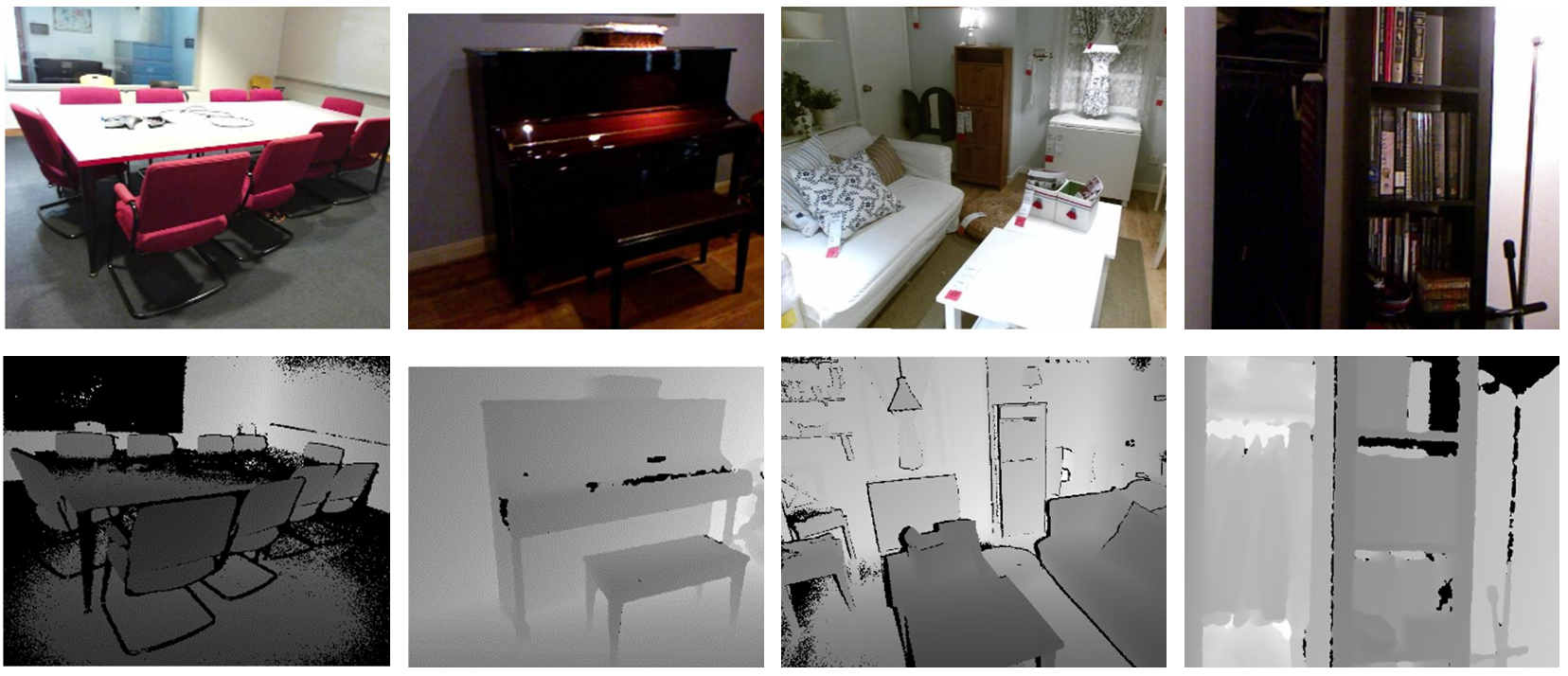}
	\end{center}
	\caption{Pairs of RGB and depth images.}
	\label{fig:rgbimages}
\end{figure*}

By introducing the Microsoft Kinect camera \cite{kinect} which captures both RGB and depth images, some indoor semantic segmentation approaches have been concentrated on the RGB-D dataset which alleviates the challenges of the indoor scene. For instance, when RGB images have a poor illumination in some regions, depth images can improve the labeling accuracy.  Figure \ref{fig:rgbimages} shows some examples in which RGB images have poor lightning in some regions while depth images hold discriminative features. 

Utilizing the 3D geometric information in semantic segmentation methods has been provided by the availability of Microsoft Kinect camera \cite{silberman2011indoor}. Extracting this 3D geometric information that might be missed in RGB images aids to diminish some uncertainties in dense prediction and object detection processes \cite{ren2012rgb, silberman2012indoor}.  Early RGB-D semantic segmentation proposed novel engineering features extracted from RGB and depth images  by using intrinsic and extrinsic camera parameters \cite{ren2012rgb, cadena2013semantic, silberman2012indoor, gupta2013perceptual, gupta2014learning}. Then, all of these appearances and 3D features were incorporated into feature vectors fed to common classifiers. 

Recently, Deep Convolutional Neural Networks (DCNNs) \cite{he2016deep, huang2017densely, simonyan2014very} have improved the accuary of almost all categories of computer vision methods; such as image classification \cite{krizhevsky2012imagenet, simonyan2014very, he2016deep}, object detection \cite{cadena2013semantic, ren2015faster}, action recognition \cite{wang2015action}, depth estimation \cite{liu2015deep, eigen2015predicting}, pose estimation \cite{kendall2015posenet} , image segmentation \cite{wang2019learning} and semantic segmentation \cite{long2015fully, chen2016deeplab, shi2018hierarchical, li2018joint}.  

Pooling operations and stride convolutions which are applied in CNNs (to become invariant to most local changes) produce the low spatial resolution outputs for dense prediction applications (such as semantic segmentation, depth estimation, and surface normal estimation). Hence, the early deep learning methods \cite{couprie2013indoor,farabet2013learning} for semantic segmentation utilized the deep networks as feature extractors. They then applied a classifier to categorize  each pixel, superpixel, or region. Long et al. \cite{long2015fully_jo} changed CNNs to Fully Convolutional Neural Networks (FCN) which are more appropriate for dense prediction applications. DeconvNet \cite{noh2015learning}, dilated convolution \cite{chen2018deeplab_j}, and unpooling \cite{badrinarayanan2017segnet_jo} methods have been proposed to recover this spatial information lost.  Among these methods, some non-parametric approaches \cite{khelifi2019mc, shuai2016scene} have been proposed where they utilize similarity measurements to label each part of images. 

As one of the goals of this paper is the semantic segmentation of RGB-Depth images, the focus is on the main challenges and approaches of RGB-D datasets. The main challenge in RGB-Depth semantic segmentation is how to represent and fuse the RGB and depth channels so that the strong correlations between the depth and photometric channels are considered. Simple methods for fusion of RGB and depth channels are based on the early fusion \cite{couprie2013indoor} and late fusion \cite{long2015fully} polices. 

\begin{figure*}
	\begin{center}
		\includegraphics[width=0.9\linewidth]{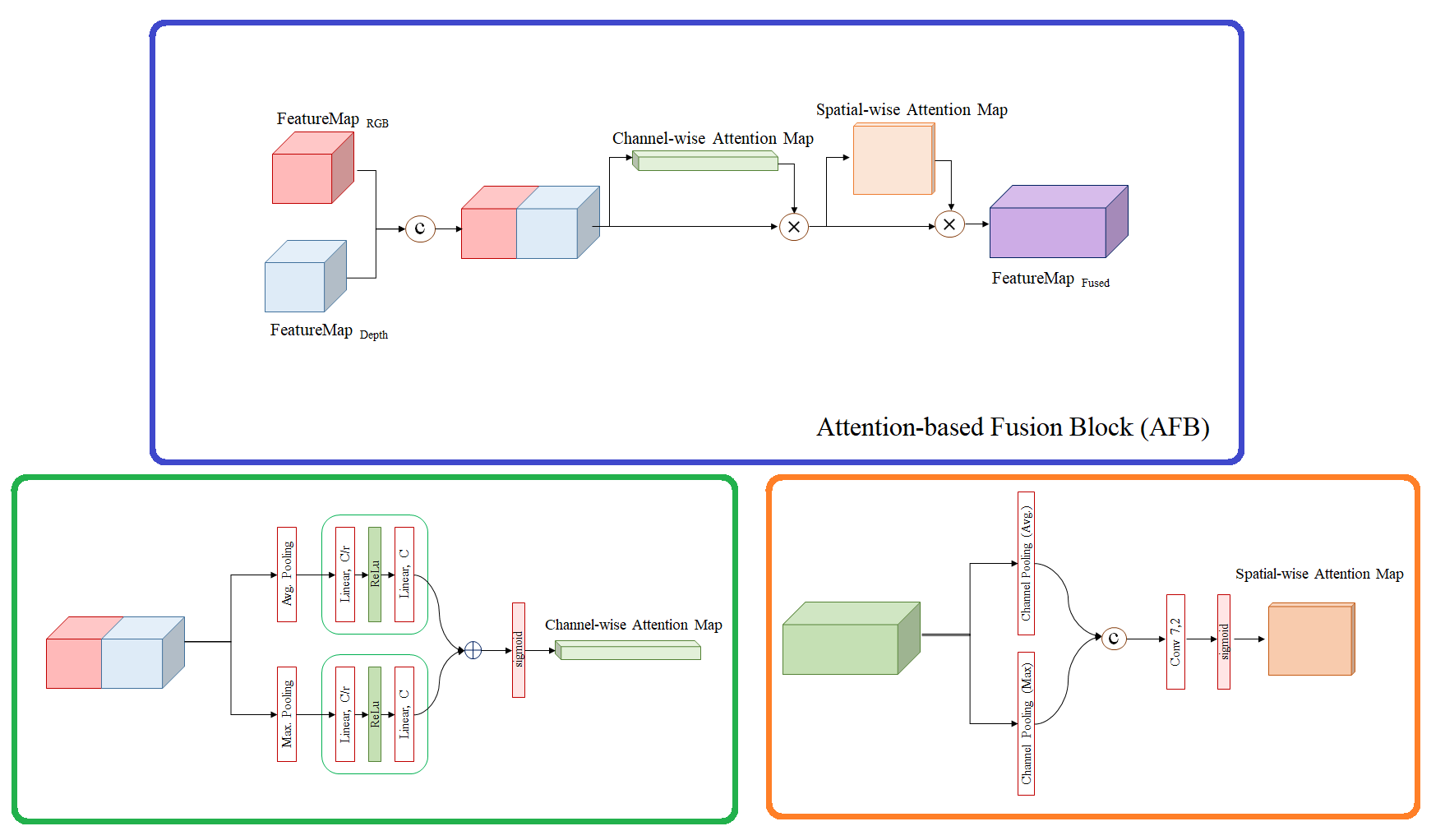}
	\end{center}
	\caption{Attention-based fusion block.}
	\label{fig:attention_block}
\end{figure*}

In this paper, the encoder-decoder architecture with the novel multi-modal Attention-based Fusion Block (AFB) is proposed to fuse these two modalities in order to obtain  more powerful and meaningful RGB-Depth fused feature maps.  The  attention-based fusion block has been inspired by the attention modules in the squeeze and excitation network \cite{hu2018squeeze} which is focused on the channel-wise recalibration of feature maps to model the dependency of channels.  The intermediate feature maps extracted from RGB and depth channels of two encoders are considered as the input to the attention-based fusion block. This block computes attention maps which are multiplied by input feature maps for adaptive feature fusion. The attention-based fusion block consists of two sequentially channel- and spatial-wise attention mechanisms to construct the attention maps. Consequently,  feature maps of two modalities are fused based on their interdependencies among different channels. Fig \ref{fig:attention_block} illustrates the proposed architecture of attention-based fusion block.  Moreover, each AFB is followed by the lightweight chained residual pooling layers to consider the global contextual information in the proposed decoder side. Consequently, the proposed encoder-decoder architecture is an efficient model in terms of the computational cost and the number of parameters.

Main contributions of this work are listed as:
\begin{itemize}
	\item Proposing an efficient encoder-decoder architecture for semantic segmentation of RGB-Depth images.  
	\item Proposing an attention mechanism of CNNs for modality fusion.
	\item Incorporating a channel-wise alongside spatial-wise interdependencies for fusion.
	\item Proposing a novel representation of evaluation metric for semantic segmentation methods.
\end{itemize}
The remainder of this paper is organized as follows. In Section \ref{sec:2}, the related work of RGB and RGB-Depth semantic segmentation are categorized. The overall architecture of proposed encoder-decoder model with the fusion block is presented in Section \ref{sec:3}. The experimental results evaluated on the existing RGB-D dataset by the proposed semantic segmentation criterion are investigated in Section \ref{sec:4}. Finally, conclusions are drawn in Section \ref{sec:5}.

\section{Related Work}
\label{sec:2}

Traditional approaches of semantic segmentation\cite{ren2012rgb, silberman2011indoor, silberman2012indoor, gupta2013perceptual, gupta2014learning, cadena2013semantic,muller2014learning,shotton2006textonboost, fooladgardicta}  have two main phases of feature extraction and classification. Engineering or hand-crafted features (such as SIFT, HOG, and SURF) are extracted from pixels, super-pixels, or segmented regions. Then, these features are fed to common classifiers; such as Support Vector Machine (SVM)  and Random Forest (RF). 

By emerging convolutional neural networks \cite{krizhevsky2012imagenet, simonyan2014very,he2016deep}, the most successful methods in the field of semantic segmentation have been proposed based on CNNs. Early CNNs methods proposed in semantic segmentation field \cite{couprie2013indoor,girshick2014rich, farabet2013learning} utilized them as a deep feature extractor.  Couprie et al. \cite{couprie2013indoor} extracted deep and dense hierarchical features for each region of the segmentation tree \cite{arbelaez2011contour} via the multi-scale CNN model. These deep features are then fed to the SVM classifier to predict the label of each region.  

Well-known CNN models (such as GoogLeNet \cite{szegedy2015going}, ResNet \cite{he2016deep,he2016identity}, and DenseNet \cite{huang2017densely}) were originally proposed for the image classification task, where the input was an image and the network output was its predicted label. But, these models need some changes to be appropriate for a dense prediction task. The cascaded down-sampling is performed by max or average pooling and then the stride convolution decrease the spatial resolution of feature maps, hence the outputs of these models for the semantic segmentation are very coarse. In other words, the localization information is lost at the end of the networks. The Fully Convolutional Network \cite{long2015fully,long2015fully_jo} converted the fully connected layers to the convolutional layers and make them suitable for a dense prediction task; such as semantic segmentation, depth estimation \cite{eigen2015predicting,liu2015deep},  surface normal prediction \cite{eigen2015predicting} and video semantic segmentation \cite{qiu2017learning}. They take an image as input and produce corresponding per pixel labeled image in an end-to-end training procedure. Fu et al. \cite{fu2018refinet} proposed Refinet model to improve the FCN method via a segmentation-based pooling idea. The goal of their pooling idea is to maintain the fine-grain details and boundary maps of salient objects.

 To recover the information loss, different approaches have been proposed. Long et al. \cite{long2015fully,long2015fully_jo} up-sampled the feature maps of the last layer  and concatenated them with the previous intermediate feature maps in a stage-wise training procedure. The encoder-decoder type models \cite{lin2017refinenet,kendall2015bayesian,badrinarayanan2015segnet} have been proposed to handle the dense per-pixel prediction problem. Commonly, the popular CNN models (such as VGG net, GoogleNet, ResNet, and DenseNet) have been utilized as the encoder to produce intermediate deep feature maps. Then, the goal of decoder branch is to restore the information lost causes in the encoder side. 
 
 Different types of approaches have been presented for the decoder side. DeconvNet \cite{noh2015learning} applied the convolution transposed in the decoder side instead of convolution layers of encoder branch. DeepLab \cite{chen2016deeplab, chen2018deeplab_j, chen2018encoder} eliminated all max-pooling layers of VGG and applied the dilated convolution to enlarge the receptive field of filters to compensate the effect of max pooling operations. Preserving the index of max-pooling and applying un-pooling operations in the decoder was presented by  SegNet model \cite{badrinarayanan2015segnet}. They proposed an encoder-decoder model in which they utilized the max-pooling indices in the decoder to recover the location of information loss. The fusion of the long-range residual connections of ResNet model was propounded by Lin et al \cite{lin2017refinenet} to refine the resolution loss of the CNN architecture. 
 
The main goal of this paper is to address the semantic segmentation challenges of RGB-Depth images. The major issue is how to extract the strong feature representations of both photometric and depth channels. In the case of RGB-Depth semantic segmentation, almost all methods exploit the depth images as another channel of the image. As such, the fusion strategy plays an important role.  These strategies can be classified into early, middle, and late fusion. Long et al. \cite{long2015fully_jo} proposed the late fusion of FCN while \cite{couprie2013indoor} utilized early fusion, and \cite{hazirbas2016fusenet} applied middle fusion of RGB and depth channels. The FuseNet \cite{hazirbas2016fusenet} considered two encoder branches, one encoder branch for the depth channel and another encoder branch for fusion of RGB and depth channels. Wang et al. \cite{wang2016learning} designed a transformation block to fuse the common and specific features of the RGB and depth channels of convolution network to bridge the convolutional and deconvolutional models. 
Li et al. \cite{li2016lstm} incorporated the vertical and horizontal Long Short-Term Memorized (LSTM) method to exploit the interior 2D global contextual relations of RGB and depth channels, separately. Then, the horizontal LSTM has been applied to their concatenated feature maps. Liu et al. in \cite{liu2018rgb} improved the HHA coding of \cite{gupta2014learning} via integrating 2D and 3D information. They then extended the VGG \cite{simonyan2014very} architecture proposed by \cite{chen2016deeplab} for RGB-D semantic segmentation. They proposed the weighted summation of RGB and depth streams of a CNN model followed by a fully connected CRF to enhance the prediction.

The RDFNet proposed by Park et al. \cite{park2017rdfnet} extended the RefineNet \cite{lin2017refinenet} for RGB-Depth images. They considered two encoder streams (RGB encoder and depth encoder), one fusion stream and one decoder stream. They utilized the cascaded refinement blocks of the RefineNet as their decoder stream. The refinement process was applied to the fusion of the RGB and depth feature maps to emend the resolution loss. The Multi-Modal Multi-Resolution RefineNet (3M2RNet) \cite{fooladgar20193m2rnet} proposed the fusion of long-range residual connections of two ResNet encoder branches with  focus on the identity mapping idea of RefineNet \cite{lin2017refinenet}. Lin et al. \cite{lin2017cascaded} proposed a context-aware receptive field based on the scene resolution to incorporate the relevant contextual information. Consequently, for each scene resolution, deep features were learned specifically in a cascaded manner to exploit the relevant information of the neighborhood. Depth-aware convolution and pooling operations were presented by  \cite{wang2018depth} to investigate the 3D geometry of the depth channel. Kang et al. \cite{kang2018depth} have proposed the depth-adaptive receptive field in the fully convolutional neural network where the size of each receptive filed has been selected based on the distance of each point from the camera. Hence, they utilize the depth images to determine the size of each filter for each neuron adaptively.

\section{Proposed Method}
\label{sec:3}
\begin{figure}
	\begin{center}
		\includegraphics[width=1\linewidth]{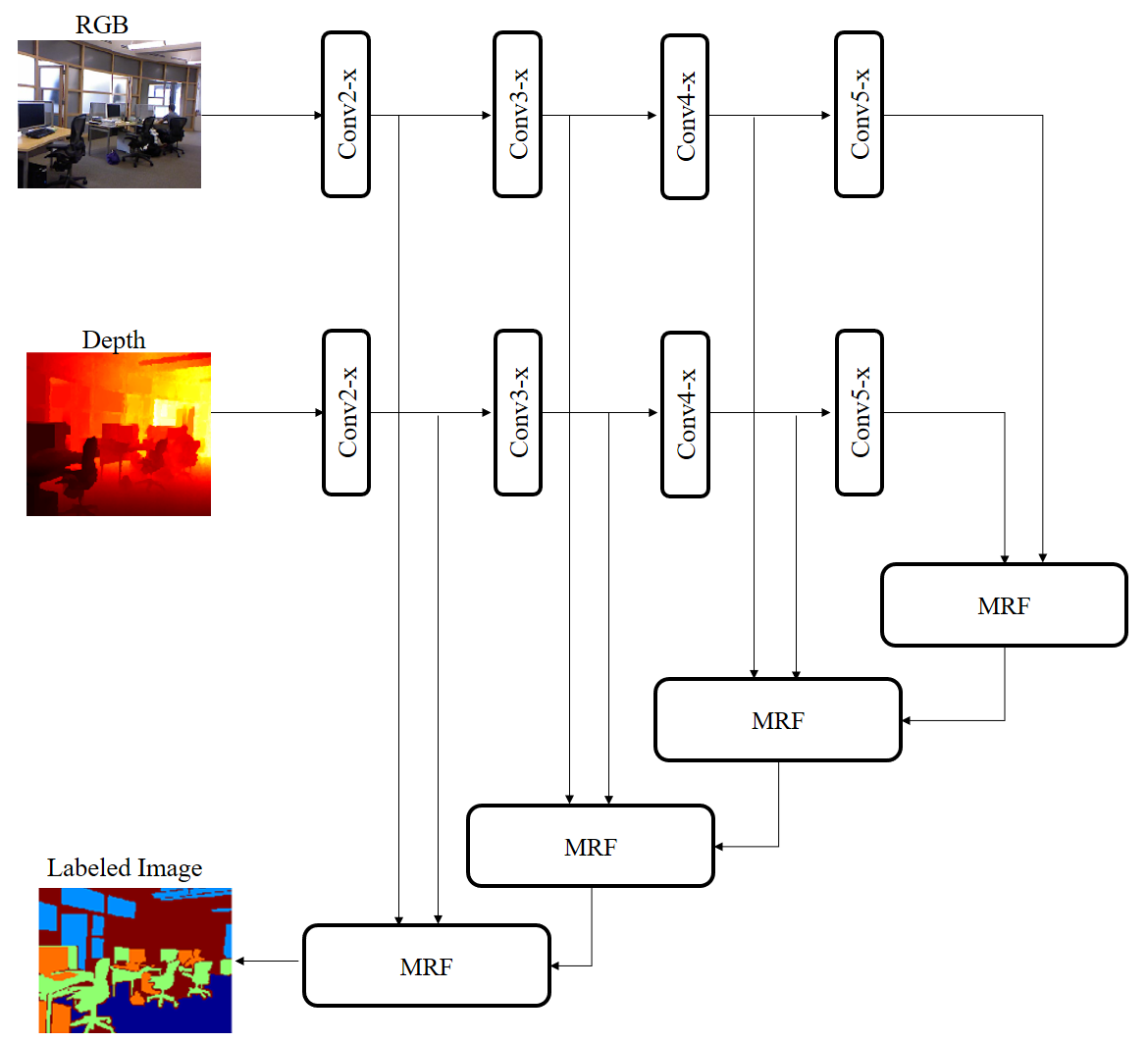}
	\end{center}
	\caption{Proposed network architecture.}
	\label{fig:CNN_model}
\end{figure}

The proposed Multi-Modal Attention Fusion Network (MMAF-Net)  model is an encoder-decoder CNN architecture with two simultaneously encoder branches of RGB and depth modalities as inputs while including one decoder branch. Both of the encoder branches follow the structure of residual block proposed in the ResNet model \cite{he2016deep}. In the decoder branch, the feature maps of both encoder branches from the same level of resolution have been fused based on the novel proposed attention fusion modules to combine both appearance and 3D feature maps. These fused feature maps have been utilized to recover the information loss of encoders and produce a high resolution prediction output. The overall view of proposed encoder-decoder architecture is illustrated in Figure \ref{fig:CNN_model}. In the following subsections, the proposed encoder-decoder architecture with a more focus on the multi-modal multi-resolution fusion block of the proposed decoder stream is explained. 

\subsection{Encoder Stream}

All of the well-known CNN models were primarily proposed for image classification. In those networks, at the end of networks the high semantic but low spatial resolution feature maps produce rough segmentation results for semantic segmentation purposes. To overcome this limitation, an encoder-decoder model is proposed. In the encoder part of the proposed model, the residual blocks of the ResNet model are utilized to benefit from the short and long range skip connections properties. The short-range skip connections immune the networks from the vanishing gradients problem while the long-range skip connections help to refine the information loss caused by the cascaded down-sampling operations and stride convolutions.

\begin{figure}
	\begin{center}
		\includegraphics[width=0.5\linewidth]{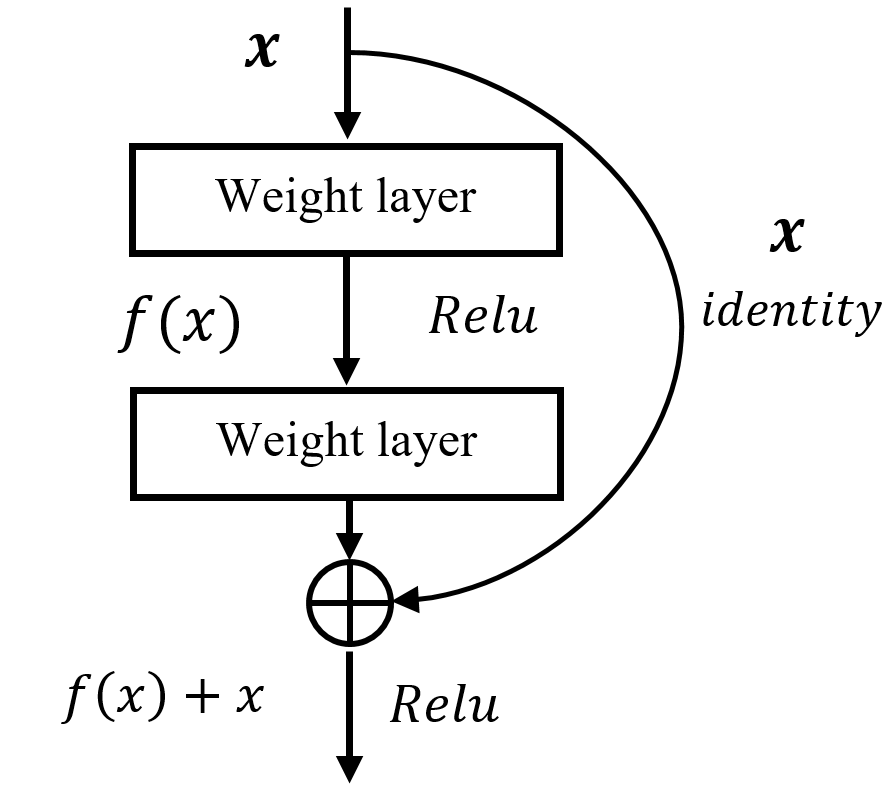}
	\end{center}
	\caption{ Building block of a residual unit\cite{he2016deep}.}
	\label{fig:resblock}
\end{figure}

As illustrated in Figure \ref{fig:CNN_model}, the proposed model utilizes the residual blocks of ResNet model (Convi-x)  as two separate encoder branches. He et al. in \cite{he2016identity}, analyzed and compared the rule of skip connections of the scaling, gating, $1\times1$ convolution, and the identity mapping. They showed that using the identity mapping function (as the skip connection) in a deep residual network is more helpful for the generalization of the network as well as the convergence of the optimization algorithm. The building block of one residual unit is illustrated in Figure \ref{fig:resblock}. This short-range skip connection can be formulated as $y_l=F(x_l,W_l )+H(x_l)$, where $x_l$ is the original input to the residual unit $l$ , and $x_{l+1}=G(y_l)$ is the output of unit $l$, which is fed to the residual unit of $l+1$ as its input. Also, $F(x_l,W_l )$, $H(x_l)$, and $G(y_l)$ are the series of the operations applied on the input $x_l$ or $y_l$ (such as convolution, batch normalization, and nonlinearity). In the first version of the ResNet model \cite{he2016deep},  $H$ is set as an identity function and the Relu function is used for $G$. Therefore, the information flow of  $x_l$ is not changed and added by $F(x_l,W_l)$. Hence, this residual unit with the identity and Relu functions enhances the performance of very deep networks as the vanishing gradient problem was solved. Consequently, if in the encoder-decoder model the weights of network are very small, the gradient of layers does  not completely vanish. Thereupon, the vanishing gradients problem does not occur in such a deep network. 

Between each residual block of the ResNet, the sequential down-sampling operations (applied by the pooling layers)  increase the receptive field of the filters to include more context and also prevent the growth in the number of training weights through the encoder stream. Therefore, they preserve efficient and tractable training. But, the network loses some valuable information. This information loss produces the low-resolution prediction in the dense per-pixel classification in which the localization of the semantic labels is more essential than the image classification applications. This means that the higher-level feature maps of deeper layers in the multi-encoders which encode the high-level semantic information and carry more object-level information suffer from the lack of localization information. Here, it is proposed to recover this information loss in the up-sampling process of the decoder branch by the attention-based fusion of the long-range residual connections of multi-encoder streams with the preceding decoder output.  Therefore, the decoder part is responsible to recover this resolution loss in cascaded multi-modal multi-resolution fusion blocks. 

\subsection{Decoder Stream}
The proposed model applies efficient multi-modal attention-based fusion modules in the decoder branch of the network to recover the information loss caused by the down-sampling processes in multi encoder streams. The goal of the decoder is to employ the multi-level feature maps coming from the long-range skip connections of two encoder branches to enhance the resolution which is lost by the down-sampling operation performed by the pooling or convolution layers (with $stride>1$).  

The output of residual blocks of encoder branches are employed as the long-range skip connections and are fed to the 4-cascaded sub-modules of the decoder, called Multi-Modal Multi-Resolution Fusion (MRF) module. As such, it actually utilizes long and short residual connections. These skip connections, along with the attention-based fusion modules, enable efficient end-to-end training of RGB-Depth encoder-decoder model as well as the efficient high-resolution prediction. 

\begin{figure*}
	\begin{center}
		\includegraphics[width=0.8\linewidth]{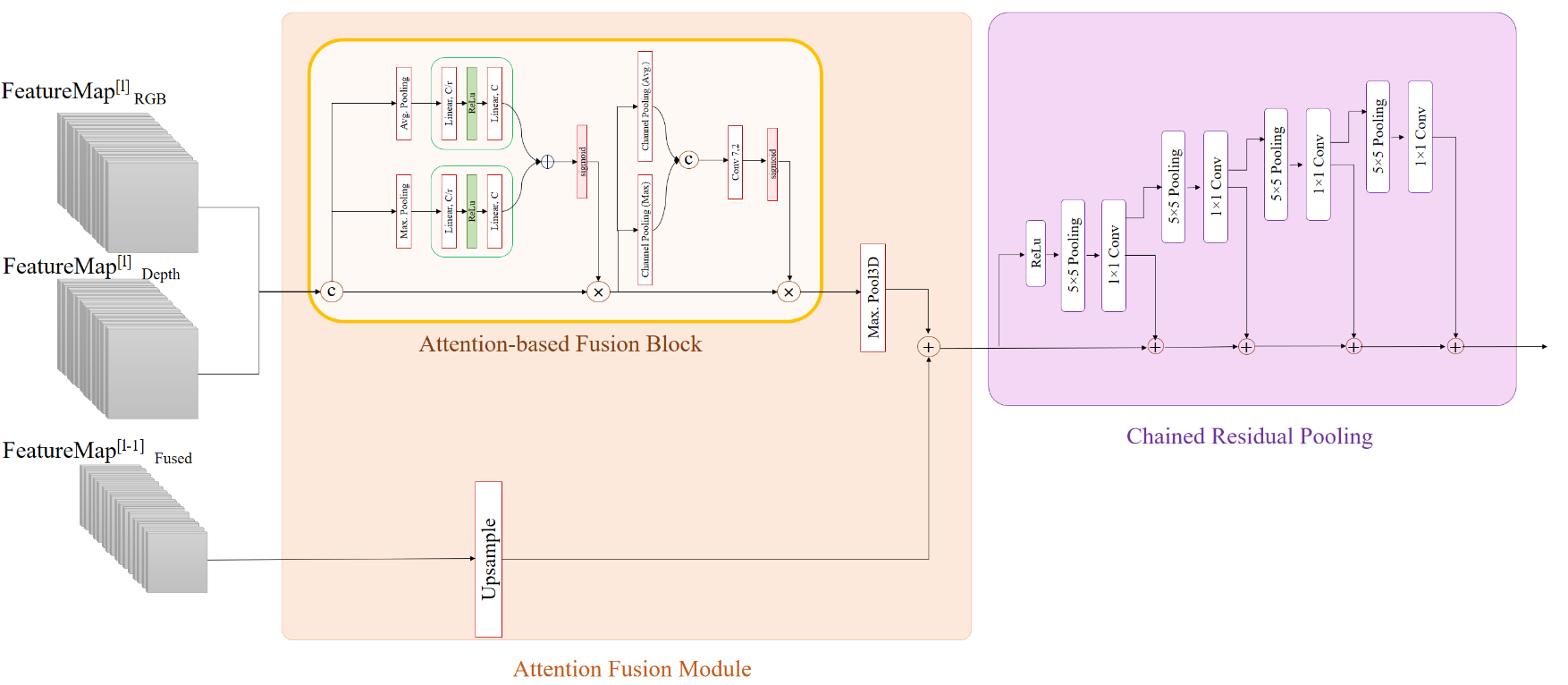}
	\end{center}
	\caption{Proposed multi-modal multi-resolution fusion module.}
	\label{fig:MMMRF}
\end{figure*}

The overall structure of the MRF module with three modalities as its inputs is illustrated in Figure \ref{fig:MMMRF}. The proposed decoder has 4 cascaded MRF modules. It consists of two main sub-blocks of Attention Fusion Module (AFM), and a Chained Residual Pooling (CRP). It has three input modalities including: i) feature maps extracted from the RGB encoder branch, ii) feature maps extracted from the depth encoder branch at the same resolution level, and iii) the feature maps of the preceding MRF at the lower resolution. In the AFM, two fusion policies have been performed. The first one is the AFB, where the attention-based fusion strategy is applied to two first inputs of this module.  The second one is a simple summation strategy to fuse the output of the previous MRF with the output of the current AFB to perform the refinement and produce the high resolution feature maps. The idea of CRP sub-block is to capture the context in multiple region sizes with the chained residual pooling layers. 

\subsubsection{Attention-based Fusion Block}
The attention mechanism in deep convolutional neural networks is based on visual attention which is consistent with the human visual perception. To perceive the scene and object structure,  human visual system focuses on the salient parts. In fact, it concentrates on the most noticeable or important parts of the scene in different sequential glances. We propose to investigate this attention for modality fusion to focus on the salient parts of feature maps in each modality. Recently, different attention-based and salient object detection \cite{li2017gla, quan2017unsupervised, fu2017saliency} have been proposed where their goal is to model the human attention mechanism. To the best of our knowledge, no previous work has investigated the attention strategy for modality fusion.

 The proposed AFB block sequentially considers channel-wise and spatial-wise attention. The goal of  channel-wise attention is to determine salient channels of concatenated feature maps, while spatial-wise attention denotes "where" salient feature maps are located. 

The proposed encoder architecture consists of four sub-blocks with the convolution, non-linearity function, and down-sampling operation. Hence, each encoder branch produces four intermediate feature maps. Intermediate feature maps of the same level in two encoder branches are concatenated and fed to the corresponding AFB in their level. 

The structure of the proposed AFB is inspired by the Convolutional Block Attention Module (CBAM) of \cite{woo2018cbam}. They proposed two sequential channel and spatial attention modules to refine the intermediate extracted feature maps. They illustrated that this module integrates the focus of the network to the target object in an image.

The goal of our attention-based fusion is to enhance the representation power of concatenated RGB-Depth feature maps and capture their salient feature maps while suppressing the unnecessary ones. The intermediate feature maps of each encoder stream demonstrate a set of local descriptors where their statistics can be considered as a good representative for each image. These statistics include the average and maximum  of each feature map. In the proposed fusion method, the non-linear and non-mutually exclusive relationships between RGB and Depth intermediate feature maps are exploited via the pooling, non-linearity, convolution, and fully connected layers of the deep neural networks operations. 

Suppose $F_{RGB}\in \mathbb{R}^{n \times m\times c}$ and $F_D \in \mathbb{R}^{n \times m\times c}$ are intermediate feature maps of RGB and Depth modalitis in the same level, respectively, and $F=[F_{RGB};F_D] \in \mathbb{R}^{n \times m\times 2c}$ shows their concatenation. The channel-wise attention map, $M_c$, is computed as

\begin{equation}
\label{eq:1}
\begin{aligned}
M_c(F) =    \sigma  (MLP(AvgPool(F)) + MLP(MaxPool(F))) \\
=   \sigma   (W_1(W_0(AvgPool(F))+W_1(W_0(MaxPool(F)))
\end{aligned}
\end{equation}

where $ \sigma$ denotes the sigmoid function. Then, $F^{'}\in \mathbb{R}^{n \times m\times 2c}$ is determined as an output of the channel-wise attention module ($F^{'} = M_c(F)\times F$). The spatial-wise attention map, $M_s$, is applied on $F^{'} $ and is  computed as

\begin{equation}
\begin{aligned}
\label{eq:2}
M_s(F^{'}) =    \sigma  (Conv([AvgPool(F^{'});MaxPool(F^{'})])).
\end{aligned}
\end{equation}

Then, $F^{''}\in \mathbb{R}^{n \times m\times 2c}$ is determined as an output of the spatial-wise attention module ($F^{''} = M_s(F^{'})\times F^{'}$).
Consequently, $F^{fused}\in \mathbb{R}^{n \times m\times c}$ is determined by an output of the spatial-wise attention module ($F^{fused} = MaxPool3D(F^{''})$).  This fused feature map focuses on the important features of the channels and concentrates on "where" salient features are located.

\subsubsection{Chained Residual Pooling}
To efficiently recover the information loss of encoder feature maps, the contextual information is exploited from large regions of an image by utilizing the cascaded pooling operations in a chain residual connections (see Figure \ref{fig:MMMRF}). The 4-cascaded $5\times5 Pooling-1\times1 Conv$ layer can capture long-range contextual information with a fixed pooling window size. These pooling feature maps combine with each other via the learnable $1\times1$ filters of convolutional layers. Consequently, all of these pooling feature maps are located in residual connections; hence each output fuses with input feature maps by a simple summation operation. 
\subsection{Training Procedures of MMAF}
To learn the proposed network, the training set is defined as
\begin{equation*}
\begin{aligned}
\label{eq:2}
\{(X_{i}^{RGB},X_{i}^{D},Y_i ) |  X_{i}^{RGB}\in \mathbb{R}^{(H\times W \times 3)}, X_i^D \in \mathbb{R}^{(H\times W)}, \\
	 Y_i  \in L^{(H\times W)}  ,i=1,…,N\}
\end{aligned}
\end{equation*}
where $L$ denotes the labeling set defined as $L={1,…,C}$ in which $C$ and $N$ determine the number of class labels and training data, respectively. The output of the networks is considered as the function $f(x^{RGB},x^D;\mathbb{W})$ that is the composition of functions corresponding to each network layer and $\mathbb{W}$ denotes all of the network weights. The probability of a pixel $x$ for a given class $c$ with the soft-max function is computed as 
\begin{equation}
\begin{aligned}
\label{eq:2}
p(\hat{y}=c |  x^{RGB},x^D  ,\mathbb{W})=\frac{e^{(f_c (x^{RGB},x^D;\mathbb{W}))}}{\sum_{l=1}^{C} e^{(f_l (x^{RGB},x^D;\mathbb{W}))}}.
\end{aligned}
\end{equation}

The simple categorical cross-entropy for the loss function is defined as

\begin{equation}
\begin{aligned}
\label{eq:2}
J = \frac{-1}{M} \sum_{i=1}^{M} \boldsymbol\ell(y_i,\hat{y_i})
\end{aligned}
\end{equation}
where $
\boldsymbol\ell(y,\hat{y}) = - \sum_{c=1}^{C} y_c \log p(\hat{y}=c | x_i^{RGB},x_i^D  ,\mathbb{W})
$.  $M$ is the total number of pixels in the training data and $y$ is the one-hot encoding vector of length $C$ that determines the ground-truth label of pixel $i$.


\section{Experimental Results}
\label{sec:4}
This section contains two main subsections. In the first one, the proposed attention-based fusion method is evaluated on the challenging  SUN-RGBD \cite{song2015sun}, NYU-V2 \cite{silberman2012indoor}, and Stanford-2D-3D-Semantic \cite{armeni2017joint} datasets. These three datasets contain RGB and depth images with the corresponding dense per pixel ground-truth (GT) images. Then, in the second subsection, the evaluation metrics of semantic segmentation are perused to find a more proper approach to analyze prediction outputs of each model on each dataset.

\subsection{Attention-based Fusion Results}
 To compare the efficiency of the proposed method, it has been compared with the state-of-the-art CNN models. It is implemented using the PyTorch library. The random cropping, scaling, and flipping operations have been utilized for data augmentation. The weights of ResNet model are employed as the pre-trained weights of two encoder branches. The categorical cross-entropy has been considered as the loss function which is optimized by the Stochastic Gradient Descent (SGD) algorithm. The global accuracy (G), the mean accuracy (M), and the Intersection-over-Union (IoU)  score have been computed for evaluation purposes.

\subsubsection{Evaluation Results on SUN-RGBD Dataset} 
SUNRGB-D dataset contains 10335  RGB-Depth images as well as dense per-pixel labeling with specific train and test set splits. The dataset has assigned each pixel into one of 37 valid classes, where one class label has been assigned to void. It is worth mentioning that the distribution of labels in this dataset is considerably unbalanced and it is remarkable that approximately 25\% of pixels in the training data have not been assigned to any of 37 valid classes and are set as the void class. 

The experiments were performed on three different ResNet models as two encoder streams. To investigate the performance of the proposed multi-modal multi-resolution fusion modules, the performance of the method is compared with the simple middle fusion strategy (SMF-Net) as well as the single modality of RGB and Depth channel. The SMF-Net is the proposed model without attention-based fusion block. In fact, its attention-based fusion block has been replaced with a simple summation operation.  As reported in Table \ref{tab:1}, the attention fusion policy has attained higher accuracies by adding less than $0.5M$ parameters. For comparison purposes, the results of the model with the single encoder branch of RGB or depth modality are also reported in Table \ref{tab:1} . 

\begin{table*}
	\centering
	\caption{Performance evaluation of proposed MMAF module on SUN-RGBD dataset.}
	\label{tab:1}
	\begin{tabular}{|c|c|c|c|c|c|c|c|}
		\hline
		Methods	&Modality&	G &	M&	IoU&	W-IoU	& No. of Parameters & GFLOPs \\
		\hline
		MMAF-Net-50 (ours)& RGB &78.4& 53.1& 42.3 & 65.9& 27.4M&32.7G\\ 
		\hline
		MMAF-Net-50 (ours) &D&	74.8&	44.7&	35.4&61.3&27.4M&32.7G\\
		\hline
		SMF-Net-50 (ours) &RGB-D&	79.0&	55.2&	43.7&67.0&	52.6M&56.7G - 464.5K\\
		\hline
		MMAF-Net-50 (ours) & RGB-D & 80.0 & 57.6 & 45.5 & 68.0&53.0M & 56.7G\\
		\hline
		MMAF-Net-101 (ours)& 	RGB-D&	80.2&58.0&	46.0& 69.0& 91.0M& 95.6G\\
		\hline
		MMAF-Net-152 (ours) & 	RGB-D&81.0	&58.2	&47.0	& 	69.6&122.3M	&134.4G\\
		\hline			
	\end{tabular}
\end{table*}

The performance of the proposed method is compared with some previous approaches that utilize RGB and depth channels.  As summarized in Table \ref{tab:2}, the proposed method has achieved a higher mean and global accuracy as well as IoU score than  other well-known methods that utilize just RGB images (such as SegNet, DeepLab, RefineNet, and some others). In comparison with the state-of-the-art methods applied on RGB-D images, the proposed fusion method attains better results than the FuseNet, LSTM-CF, and D-CNN while it achieves comparable accuracy with the RDFNet and 3M2RNet methods.

The proposed multi-modal multi-resolution fusion method is more computationally efficient than these two methods. In Table \ref{tab:3}, the proposed method has been compared with them based on computational complexity and model size. The proposed method is more advisable for applications running on embedded devices or even those that require real-time performance. The RDFNet model did not report its computational cost and model size. But, it is an extension of the RefineNet for RGB-Depth images. Hence, it has inevitably more parameters and computational cost than the RefineNet, because it has one extra encoder stream for depth channel and one additional fusion stream to fuse RGB and depth feature maps. It is also notable that the accuracy of RefineNet, RDFNet, and 3M2RNet model have been reported based on a multi-scale evaluation  where all of the other accuracies are reported on a single-scale evaluation.

\begin{figure*}
	\subfloat[RGB]{\hspace{2.2cm}} \subfloat[Depth] {\hspace{2.2cm}}\subfloat[GT]{\hspace{2.2cm}}\subfloat[Prediction]{\hspace{2.2cm}}\subfloat[RGB]{\hspace{2.4cm}} \subfloat[Depth] {\hspace{2.2cm}}\subfloat[GT]{\hspace{2.4cm}}\subfloat[Prediction]{\hspace{2.2cm}}\vspace{0.1cm}
	\subfloat{\includegraphics[width=0.12\linewidth]{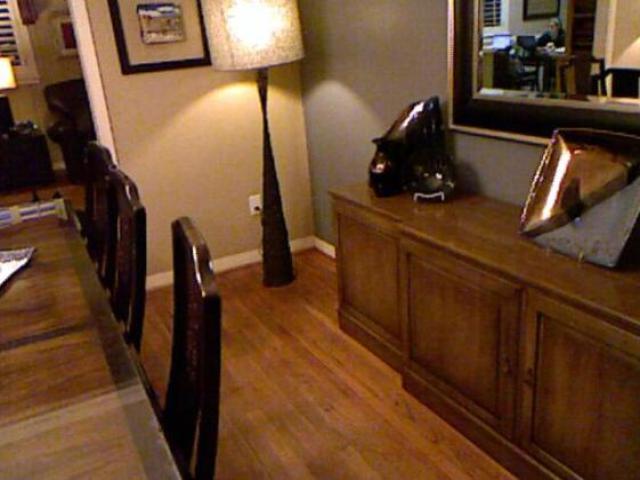}} \hspace{0.001cm}
	\subfloat{\includegraphics[width=0.12\linewidth]{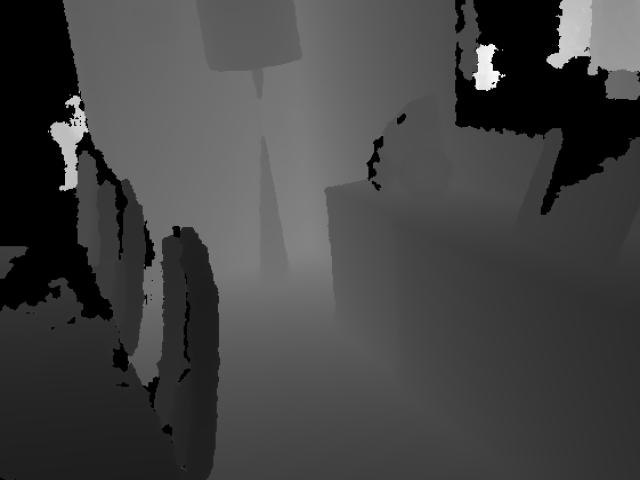}}\hspace{0.001cm}
	\subfloat{\includegraphics[width=0.12\linewidth]{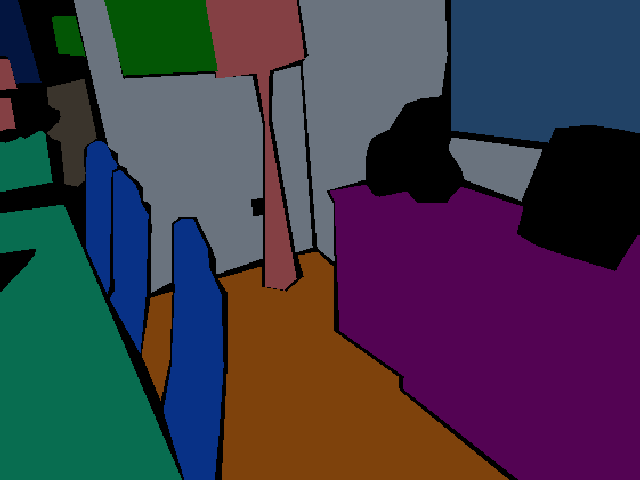}}\hspace{0.001cm}
	\subfloat{\includegraphics[width=0.12\linewidth]{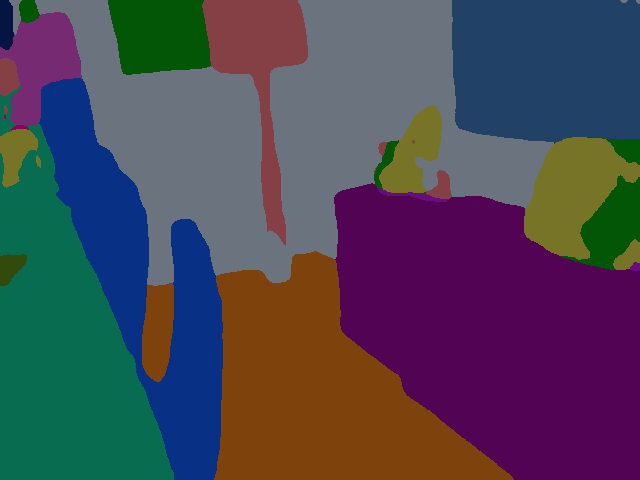}} \hspace{0.01cm}
	\subfloat{\includegraphics[width=0.12\linewidth]{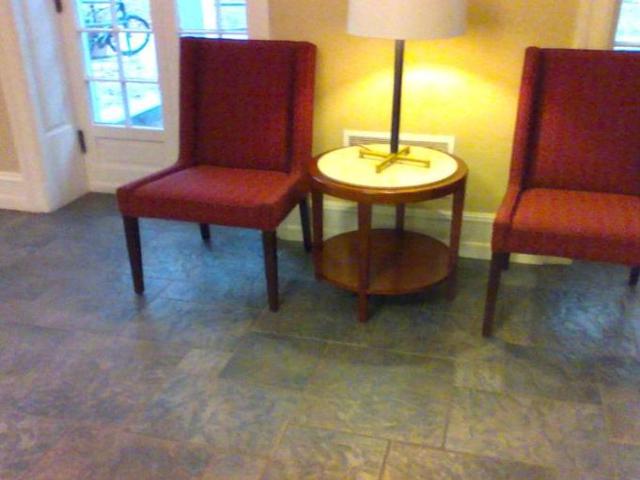}}  \hspace{0.001cm}
	\subfloat{\includegraphics[width=0.12\linewidth]{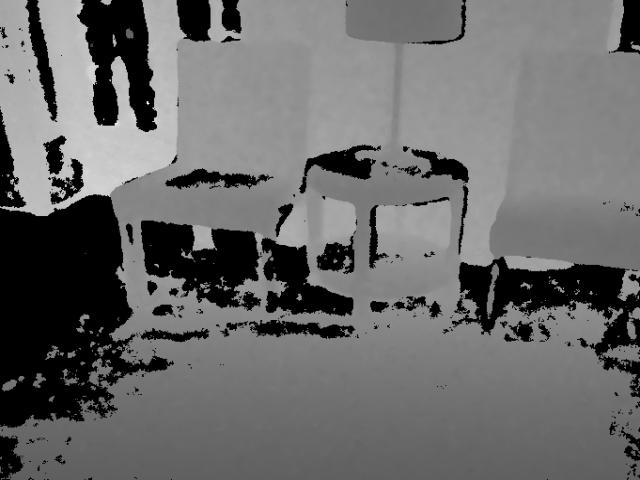}} \hspace{0.001cm}
	\subfloat{\includegraphics[width=0.12\linewidth]{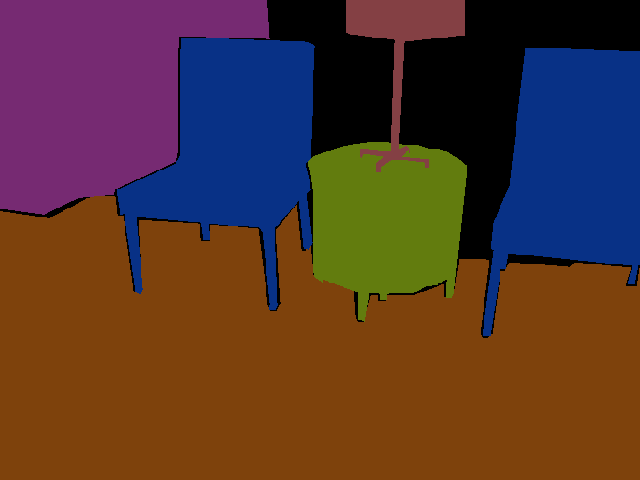}} \hspace{0.001cm}
	\subfloat{\includegraphics[width=0.12\linewidth]{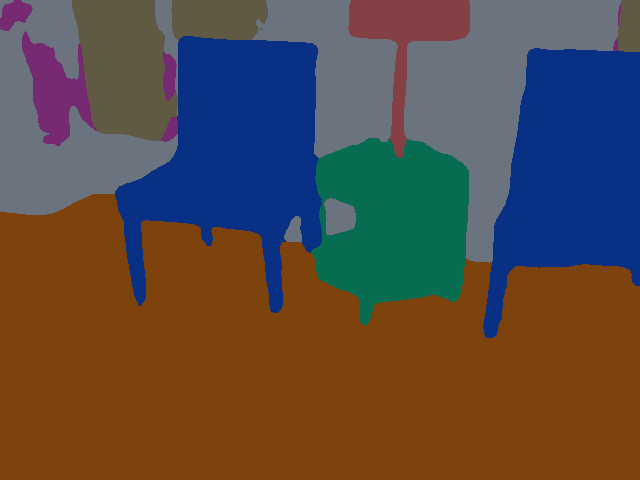}}  \hspace{0.01cm}
	\subfloat{\includegraphics[width=0.12\linewidth]{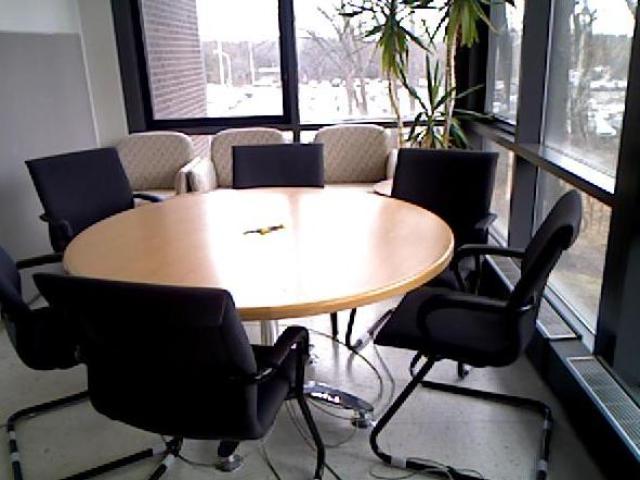}}  \hspace{0.001cm}
	\subfloat{\includegraphics[width=0.12\linewidth]{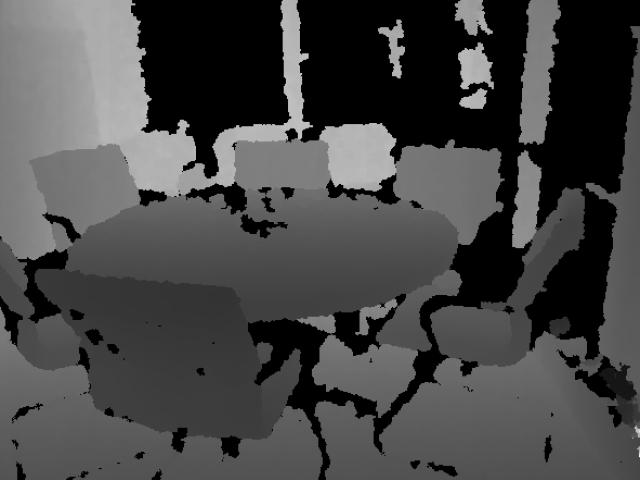}} \hspace{0.001cm}
	\subfloat{\includegraphics[width=0.12\linewidth]{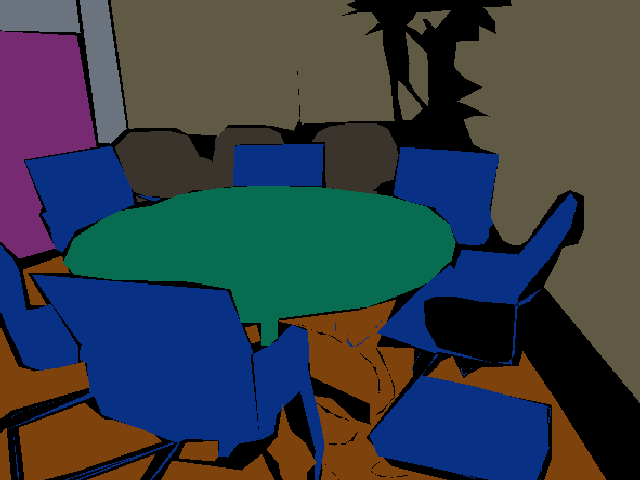}} \hspace{0.001cm}
	\subfloat{\includegraphics[width=0.12\linewidth]{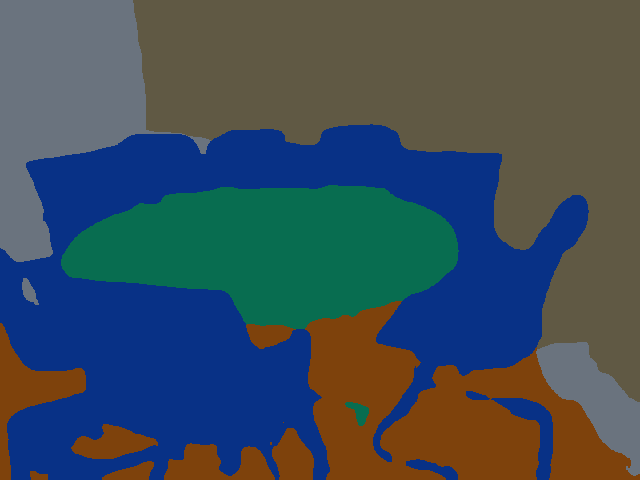}}  \hspace{0.01cm}
	\subfloat{\includegraphics[width=0.12\linewidth]{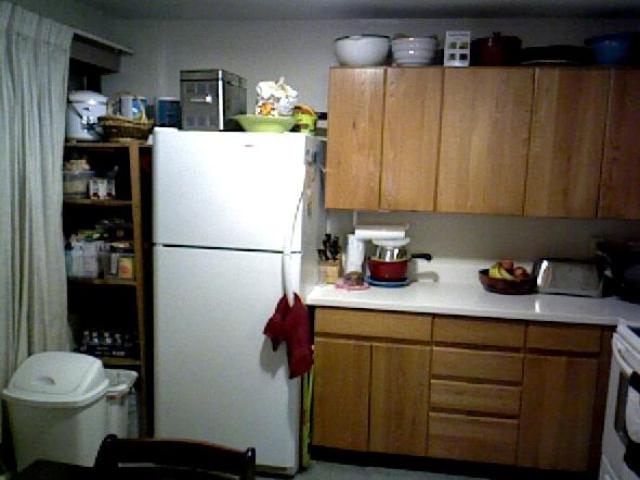}}  \hspace{0.001cm}
	\subfloat{\includegraphics[width=0.12\linewidth]{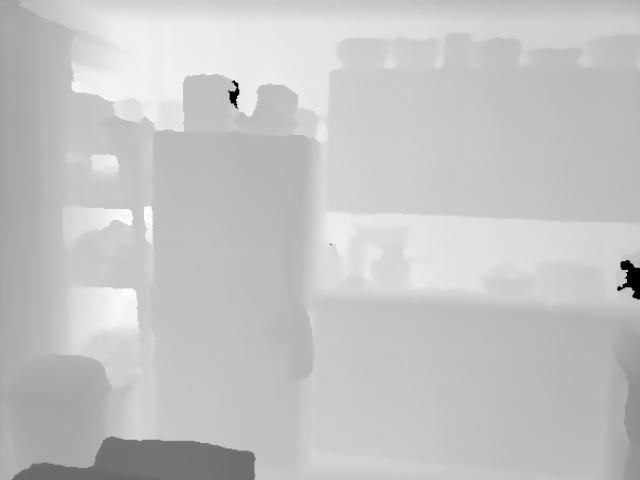}} \hspace{0.001cm}
	\subfloat{\includegraphics[width=0.12\linewidth]{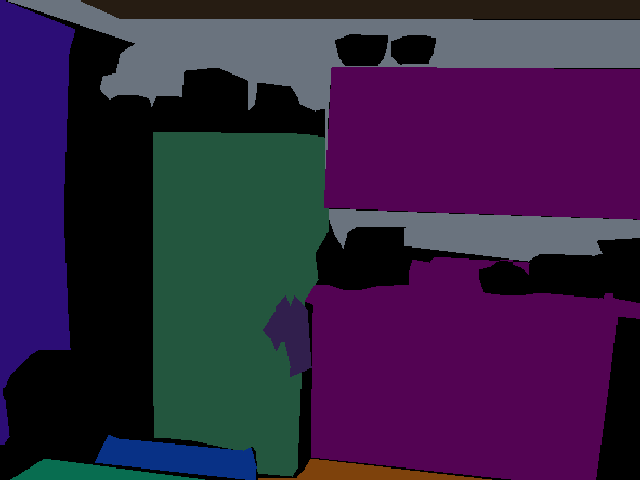}} \hspace{0.001cm}
	\subfloat{\includegraphics[width=0.12\linewidth]{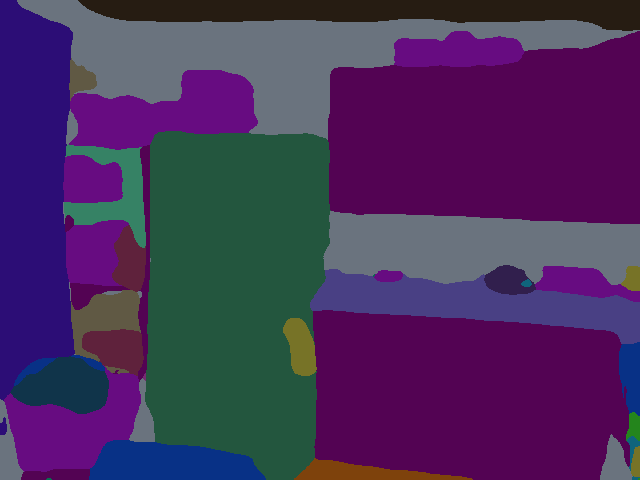}}  \hspace{0.01cm}
	\subfloat{\includegraphics[width=0.12\linewidth]{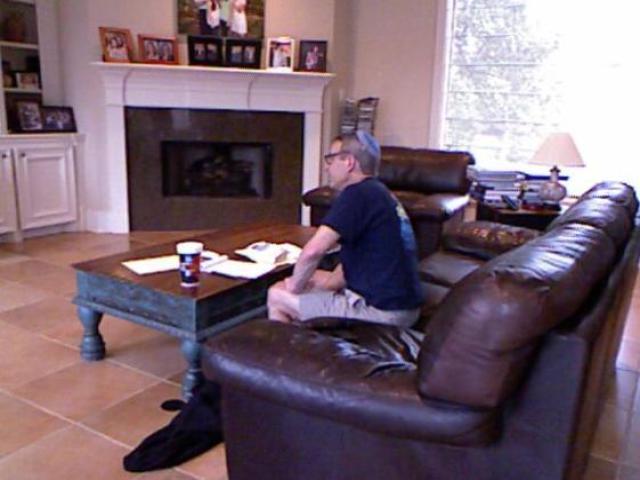}}  \hspace{0.001cm}
	\subfloat{\includegraphics[width=0.12\linewidth]{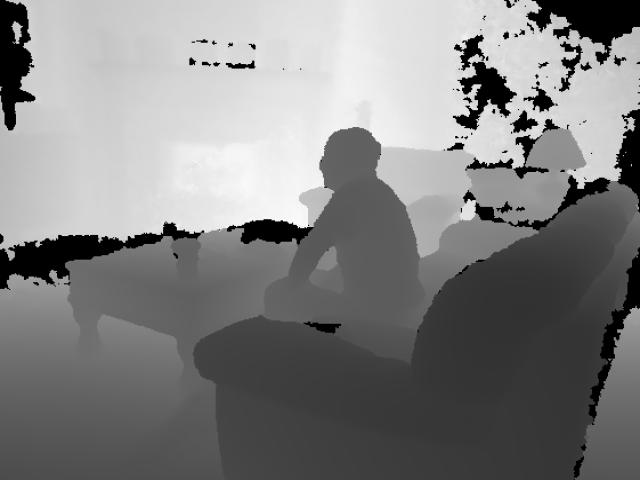}} \hspace{0.001cm}
	\subfloat{\includegraphics[width=0.12\linewidth]{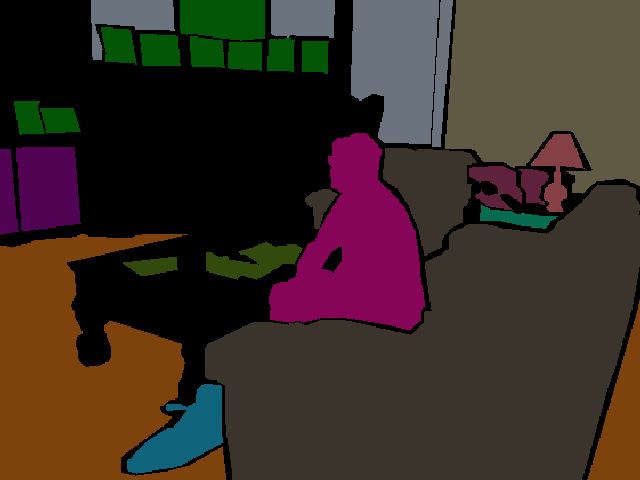}} \hspace{0.001cm}
	\subfloat{\includegraphics[width=0.12\linewidth]{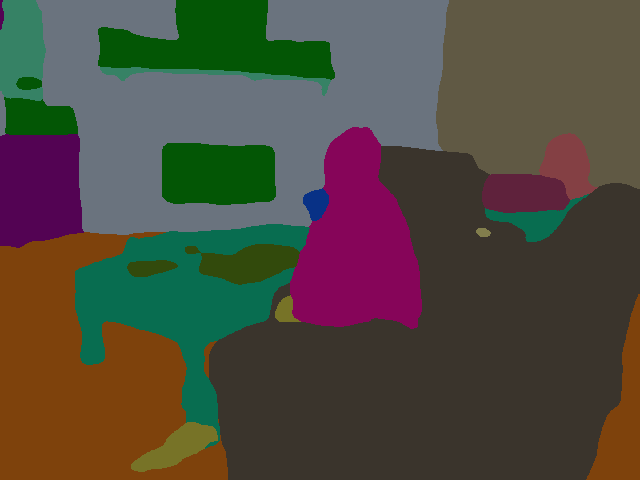}}  \hspace{0.01cm}
	\subfloat{\includegraphics[width=0.12\linewidth]{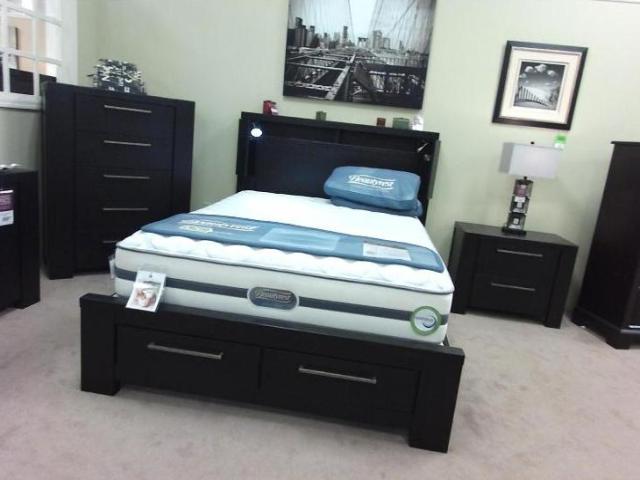}}  \hspace{0.001cm}
	\subfloat{\includegraphics[width=0.12\linewidth]{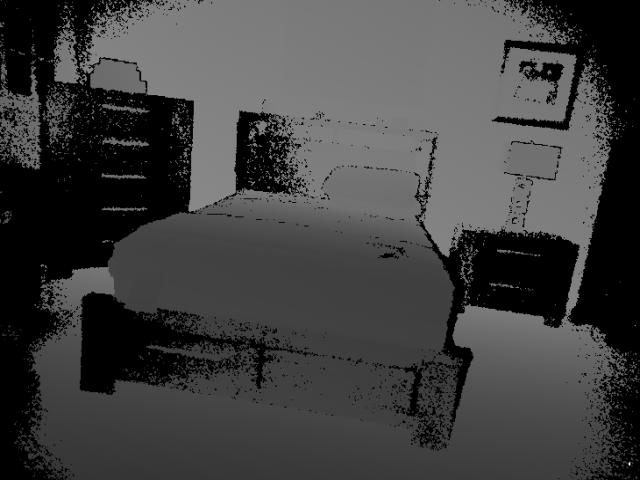}} \hspace{0.001cm}
	\subfloat{\includegraphics[width=0.12\linewidth]{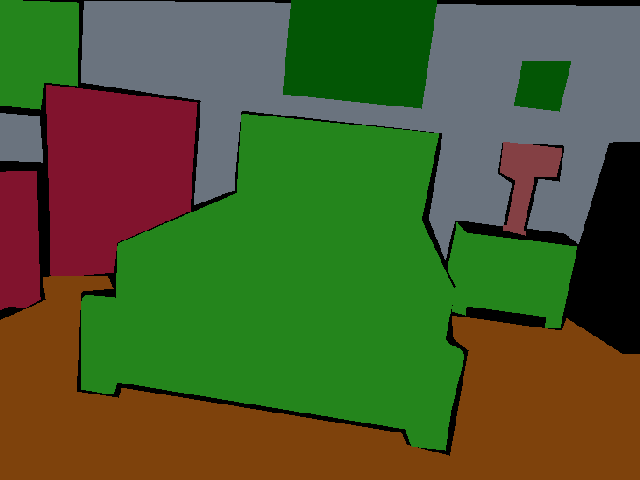}} \hspace{0.001cm}
	\subfloat{\includegraphics[width=0.12\linewidth]{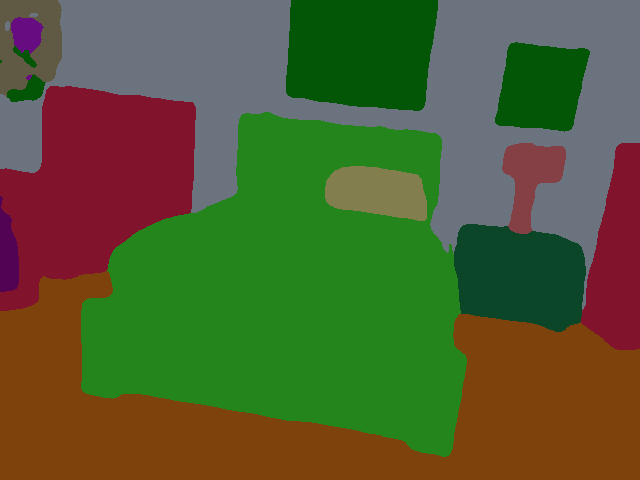}}  \hspace{0.01cm}
	\subfloat{\includegraphics[width=1\linewidth]{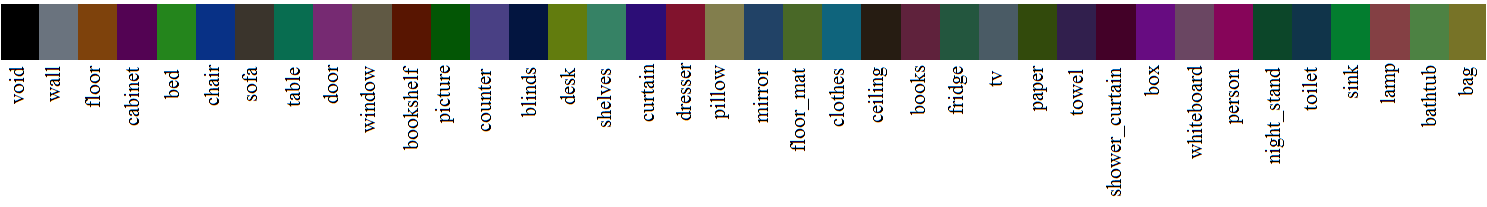}}  \vspace{0.1cm}
	\caption{Qualitative assessments of proposed method on SUN RGB-D dataset.}
	\label{fig:sun_sampel_results}
\end{figure*}

In Figure \ref{fig:sun_sampel_results}, some test samples of the SUN-RGBD dataset that are predicted by the proposed method are depicted. 
\begin{table}
	\caption{Semantic segmentation results on SUN RGB-D dataset ('$^*$' denotes multi-scale evalution).}
	\label{tab:2}
	\begin{tabular}{|c|c|c|c|c|}
		\hline
		Methods	&Modality&	G &	M&	IoU \\
		\hline
		Ren et al. \cite{ren2012rgb}&RGB-D&	-&	36.3&	-\\
		\hline
		DeconvNet \cite{noh2015learning}&	RGB	&66.1&	32.3&	22.6\\
		\hline
		FCN  \cite{long2015fully}	& RGB &	68.2&	38.4&	27.4\\
		\hline
		SegNet \cite{badrinarayanan2017segnet_jo}&	RGB &	72.6&	44.8&	31.8\\
		\hline
		B-SegNet \cite{kendall2015bayesian}&	RGB &	71.2&	45.9&	30.7\\
		\hline
		DeepLab \cite{chen2018deeplab_j}&RGB	 &71.9 &	42.2 &	32.1 \\
		\hline
		LSTM-CF \cite{li2016lstm}&	RGB-D &	-&	48.1 &	-\\
		\hline
		FuseNet \cite{hazirbas2016fusenet}&	RGB-D	&76.3&	48.3&	37.3\\
		\hline
		Context \cite{lin2018exploring}&	RGB	&78.4&	53.4&	42.3 \\
		\hline
		D-CNN \cite{wang2018depth} &	RGB-D&	-&	53.5&	42.0\\
		\hline
		3D Graph \cite{qi20173d} & RGB-D & - & 57.0 & 45.9  \\
		\hline
		Cheng et al. \cite{cheng2017localitysensitive} & RGB-D & -& 58.0 & -  \\ 
		\hline
		RefineNet \cite{lin2017refinenet} &RGB&	80.6$^*$&	58.5$^*$&	45.9$^*$\\
		\hline
		CFN (VGG-16)\cite{lin2017cascaded} & RGB-D & - & - & 42.5$^*$ \\
		\hline
		CFN (RefineNet)\cite{lin2017cascaded} & RGB-D & - & - & 48.1$^*$ \\
		\hline
		RDFNet \cite{park2017rdfnet}& 	RGB-D&	81.5$^*$&	60.1$^*$&	47.7$^*$\\
		\hline
		3M2RNet \cite{fooladgar20193m2rnet}& 	RGB-D&	83.1$^*$&	63.5$^*$&	49.8$^*$\\
		\hline
		MMAF-Net-152 (ours) & 	RGB-D&81.0	&58.2	&47.0	\\
		\hline			
	\end{tabular}
\end{table}

\begin{table}
	\centering
	\caption{Computational complexity and model size comparison on SUN RGB-D dataset ('$^*$' denotes multi-scale evalution).}
	\label{tab:3}
	\begin{tabular}{|c|c|c|c|c|c|}
		\hline
		Methods	&	G &	M&	IoU	& Parameters & GFLOPs \\
		\hline
		RefineNet \cite{lin2017refinenet}&80.6$^*$&	58.5$^*$&	45.9$^*$&119.0M & 234.9G\\ 
		\hline
		3M2RNET \cite{fooladgar20193m2rnet} &		83.1$^*$&	63.5$^*$&	49.8$^*$&225.4M&384.5G\\
		\hline
		RDFNET \cite{park2017rdfnet} &81.5$^*$&	60.1$^*$&47.7$^*$&	-&-\\
		\hline
		MMAF-Net (ours) &81.0	&58.2	&47.0	& 122.3M&134.4G\\
		\hline			
	\end{tabular}
\end{table}

\subsubsection{Evaluation Results on NYU-V2 Dataset} 
The NYU-V2 dataset is known as the most popular dataset among indoor RGB-D datasets. It contains images from 646 different scenes with 26 variants  of scene types. It includes 1449 RGB and depth images with per-pixel annotation which are splitted to 795 training images and 654 test images. Their class labels are mapped to 40 class labels by Gupta et al. in \cite{gupta2013perceptual}. The dataset is unbalanced with respect to the ratio of the number of pixels per class objects and contains the label “void” showing the pixels which cannot be annotated. 

The proposed method has been compared with the most important CNN models.  As the results listed in Table \ref{tab:4}  show, it has surpassed all of CNN models using a single RGB modality. For instance, it obtained approximately $\%6$  higher mean IoU than the Context model  \cite{lin2018exploring}. It has also achieved better results than methods that have utilized both RGB and depth channels, while the MMAF-Net did not outperform the RDFNet and 3M2RNet model in terms of accuracy. But, note that it has a lower model size as well as computational complexity than these two models (see Table \ref{tab:3}). These two models, as well as the RefineNet, have evaluated their results based on the multi-scale method (determines with '$^*$' in Table \ref{tab:4}).

\begin{table}
	\caption{Semantic segmentation results on NYU-V2 dataset ('$^*$' denotes multi-scale evalution).}
	\label{tab:4}
	\begin{tabular}{|c|c|c|c|c|}
		\hline
		Methods	&	Modality &	G.&	M&	IoU	\\
		\hline
		Silberman et al. \cite{silberman2012indoor}&	RGB-D&	54.6&	19.0&	- 	\\
		\hline
		Ren et al. \cite{ren2012rgb}&	RGB-D&	49.3&	21.1&	21.4 \\
		\hline
		Gupta et al. \cite{gupta2013perceptual}& RGB-D&	59.1&	28.4&	29.1\\
		\hline
		Gupta et al. \cite{gupta2014learning}&	RGB-D&	60.3&	35.1&	31.3\\
		\hline
		Eigen et al. \cite{eigen2015predicting}&	RGB&	65.6&	45.1&	34.1\\
		\hline
		FCN \cite{long2015fully}&	RGB-D&	65.4&	46.1&	34.0\\
		\hline
		Wang et al. \cite{wang2016learning}	&RGB-D&	-&	47.3&	-\\
		\hline
		Liu et al. \cite{liu2018rgb} &	RGB-D&70.3& 51.7& 41.2\\
		\hline
		Context \cite{lin2018exploring}&	RGB&	70.0&	53.6&	40.6\\
		\hline
    	Kang et al. \cite{kang2018depth}& RGB-D&68.4&49.0&37.6\\
    	\hline
		LSTM-CF \cite{li2016lstm}&	RGB-D	& -	& 49.4 &	- \\
		\hline
		3D Graph \cite{qi20173d} &RGB-D & - &55.7 & 43.1  \\
		\hline
		D-CNN \cite{wang2018depth} & RGB-D&	-&	56.3&	43.9\\
		\hline
		Cheng et al. \cite{cheng2017localitysensitive} & RGB-D& 71.9 & 60.0 & 45.9\\
		\hline
		RefineNet \cite{lin2017refinenet}&		RGB&	73.6$^*$&	58.9$^*$&	46.5$^*$\\
		\hline
		CFN (VGG-16) \cite{lin2017cascaded}& RGB-D& -&-&41.7$^*$\\
		\hline
		CFN (RefineNet)\cite{lin2017cascaded}& RGB-D& -&-&47.7$^*$\\
		\hline
		RDFNet \cite{park2017rdfnet}&	RGB-D&	76.0$^*$&	62.8$^*$&	50.1$^*$\\
		\hline
		3M2RNet \cite{fooladgar20193m2rnet} & 	RGB-D&	76.0$^*$&	63.0 $^*$&	48.0$^*$\\
		\hline
		MMAF-Net-152 (ours) & 	RGB-D&72.2	&59.2	&44.8\\
		\hline		
	\end{tabular}
\end{table}

\subsubsection{Evaluation Results on Stanford-2D-3D-Semantic Dataset} 
It contains 70496 RGB and depth images as well as 2D annotation with 13 object categories. It includes 1413 RGB and depth panoramic images  as well as their surface normal and semantic annotations of six large-scale indoor areas. It also provides 3D point clouds of these areas. Areas 1, 2, 3, 4, and 6 are utilized as the training and Area 5 is used as the testing set. The attention-based fusion model was applied on RGB-D images (not panoramic ones). Hence, Table \ref{tab:5} shows the performance comparison with those approaches that have been evaluated on RGB-D images. The authors of the D-CNN model \cite{wang2018depth}, evaluated their model as well as the DeepLab \cite{chen2016deeplab} model on this dataset. They trained these two models from scratch. The proposed MMAF-Net has obtained comparable performance with the 3M2RNet model in terms of accuracy while it enjoys a lower model size and computational complexity (see Table \ref{tab:3}). Tateno et al. \cite{tateno2018distortion} and Kong et al. \cite{kong2018pixel} reported their accuracy on the panoramic images of this dataset.
\begin{table}
	\caption{Semantic segmentation results on Stanford-2D-3D-Semantic dataset ('$^*$' denotes multi-scale evalution).}
	\label{tab:5}
	\begin{tabular}{|c|c|c|c|c|}
		\hline
		Methods	&	Modality &	G.&	M&	IoU	\\
		\hline
		 DeepLab \cite{chen2016deeplab} & RGB-D&	64.3&	46.7&	35.5\\
		\hline
		 D-CNN \cite{wang2018depth} & RGB-D&65.4&	55.5&	39.5\\
		\hline
		3M2RNet \cite{fooladgar20193m2rnet} & 	RGB-D&	79.8$^*$&	75.2 $^*$&	63.0$^*$\\
		\hline
		MMAF-Net-152 (ours) & 	RGB-D&76.5	&62.3	&52.9\\
		\hline		
	\end{tabular}
\end{table}

\subsection{Proposed Evaluation Metrics for Semantic Segmentation }
The semantic segmentation problem is actually known as a dense labeling problem. Hence, evaluation metrics that had been utilized for labeling in the machine learning field have also been applied to semantic segmentation methods. Accordingly, the confusion matrix has been computed and then the global, mean, and IoU criteria have been figured out from it. There are two main issues related to these criteria used for semantic segmentation methods which are explained in the following with more details. 

\begin{figure}
	\begin{center}
		\includegraphics[width=0.9\linewidth]{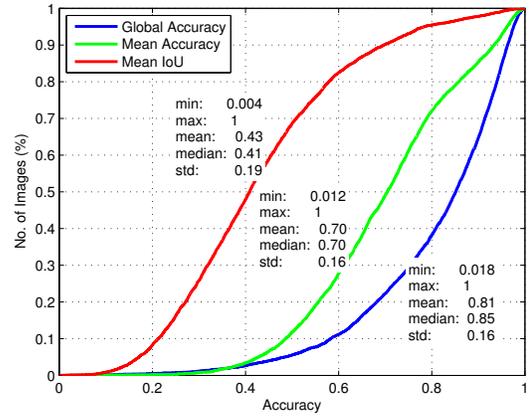}
	\end{center}
	\caption{Semantic segmentation criteria computed per image presented via CDF.}
	\label{fig:acc_per_image}
\end{figure}

\textbf{First issue:} Almost all of the semantic segmentation approaches calculate these metrics per pixel for whole images of each dataset (per dataset). For example, the $global\ accuracy = 81$ means $\%81$ of all “pixels” of all test images have been classified correctly.   It does not carry out any additional information about each image's accuracy. Hence, all pixels of one test image may be classified correctly but the other ones may have a large misclassification error. As a result, it is ambiguous whether the method has approximately the same performance for all images or it has a rich performance for some of them and a poor performance for others. Csurka et al. \cite{csurka2013good} proposed to measure the per image accuracy instead of per dataset. They computed the confusion matrix for each image based on the union of classes presented in the ground-truth as well as in the prediction. Therefore, the number of images that have attained an accuracy more than a specific threshold can be reported. But, almost all of the existing methods followed the former accuracy metrics which are computed per dataset. We propose to compute the global, mean, and IoU metrics for each test image, separately and depict their Cumulative Distribution Function (CDF) to illustrate the least number of images with a specified accuracy level. Figure \ref{fig:acc_per_image} shows this CDF for these three famous criteria. The horizontal axis shows the accuracy. Hence, for instance, approximately 70 percent of test images have more than $\%80$ global accuracy. For each CDF, minimum, maximum, median, mean, and standard deviation have been reported.

\begin{figure*}
	\subfloat{\includegraphics[width=0.33\linewidth]{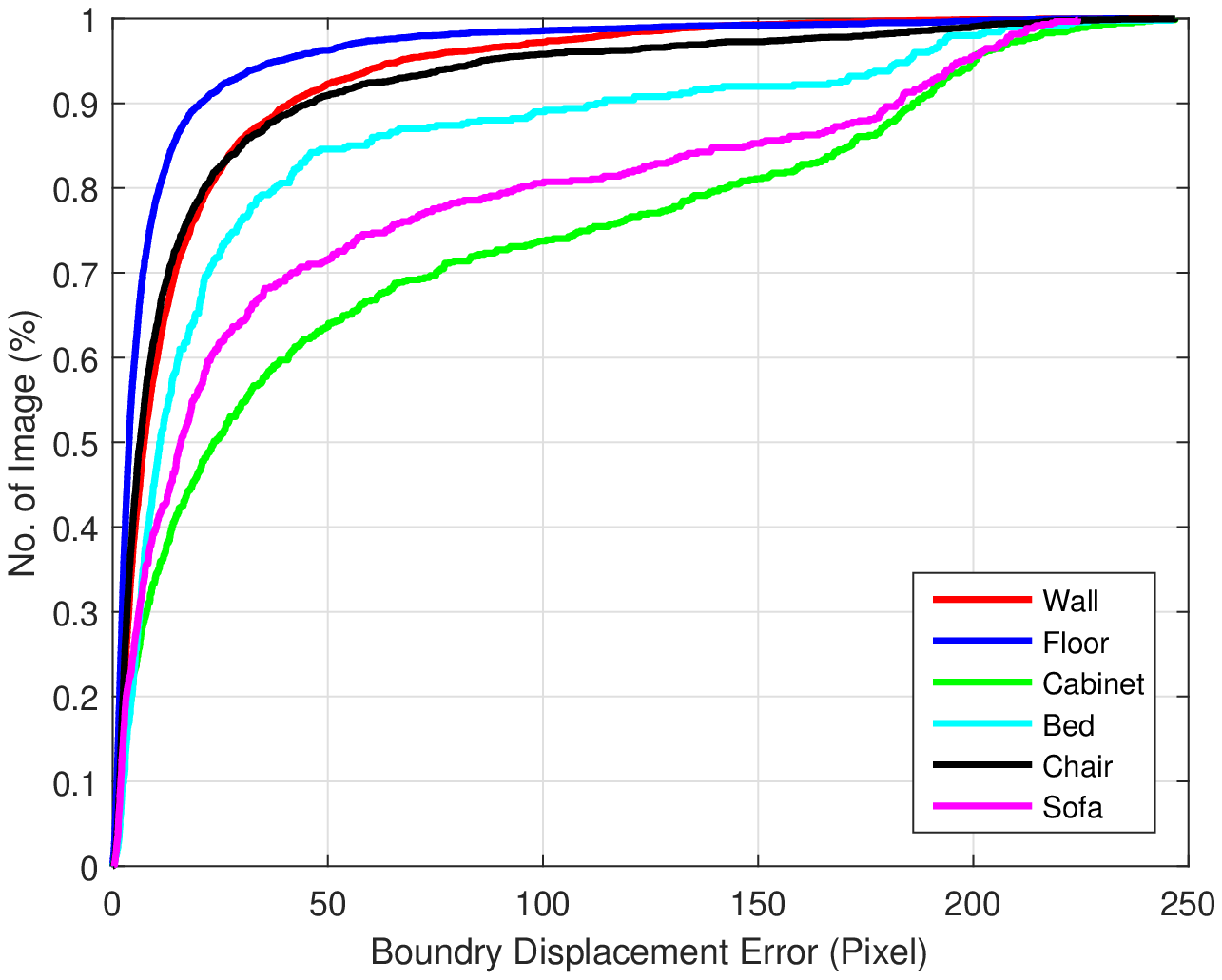}} \hspace{0.05cm}
	\subfloat{\includegraphics[width=0.33\linewidth]{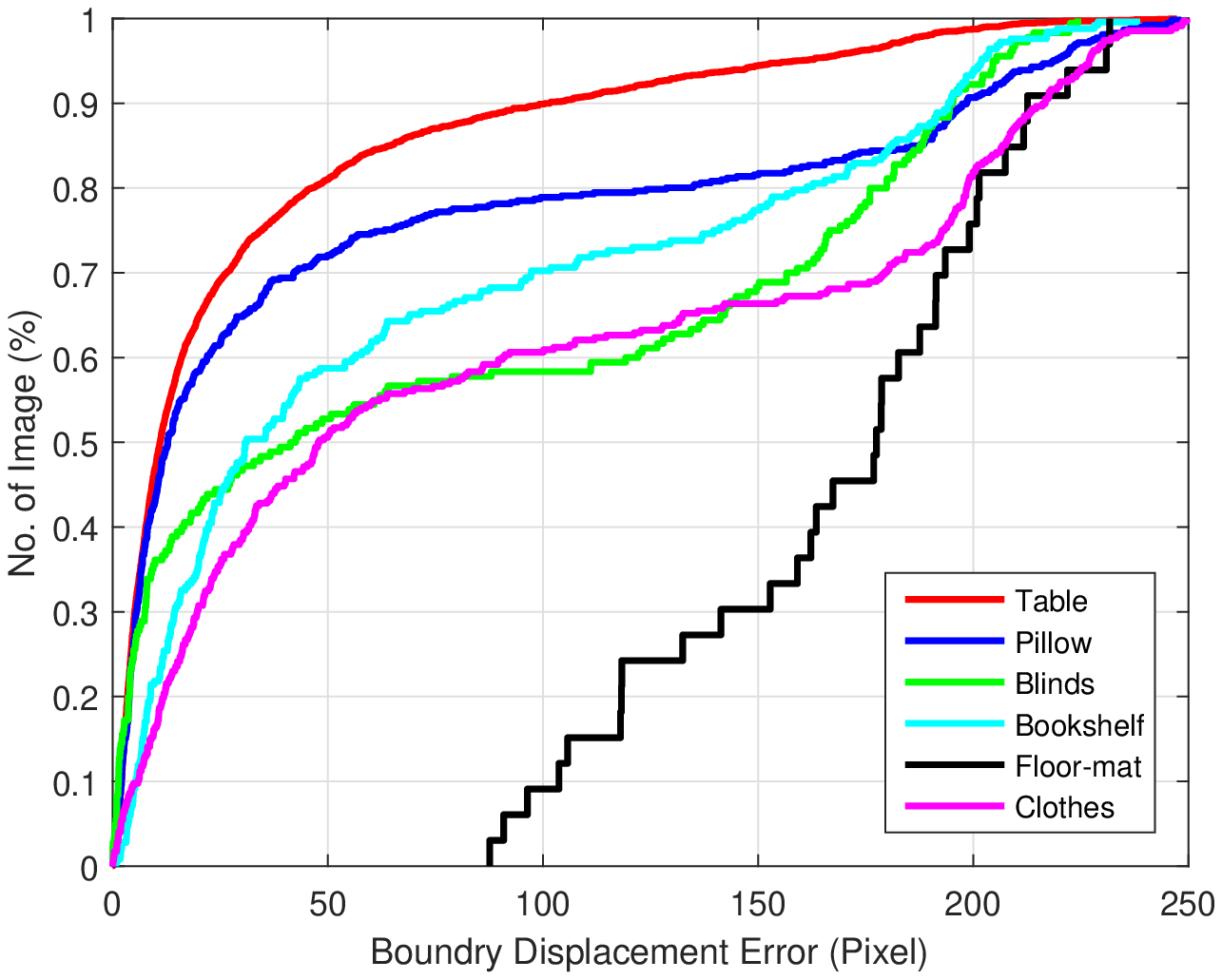}}\hspace{0.05cm}
	\subfloat{\includegraphics[width=0.33\linewidth]{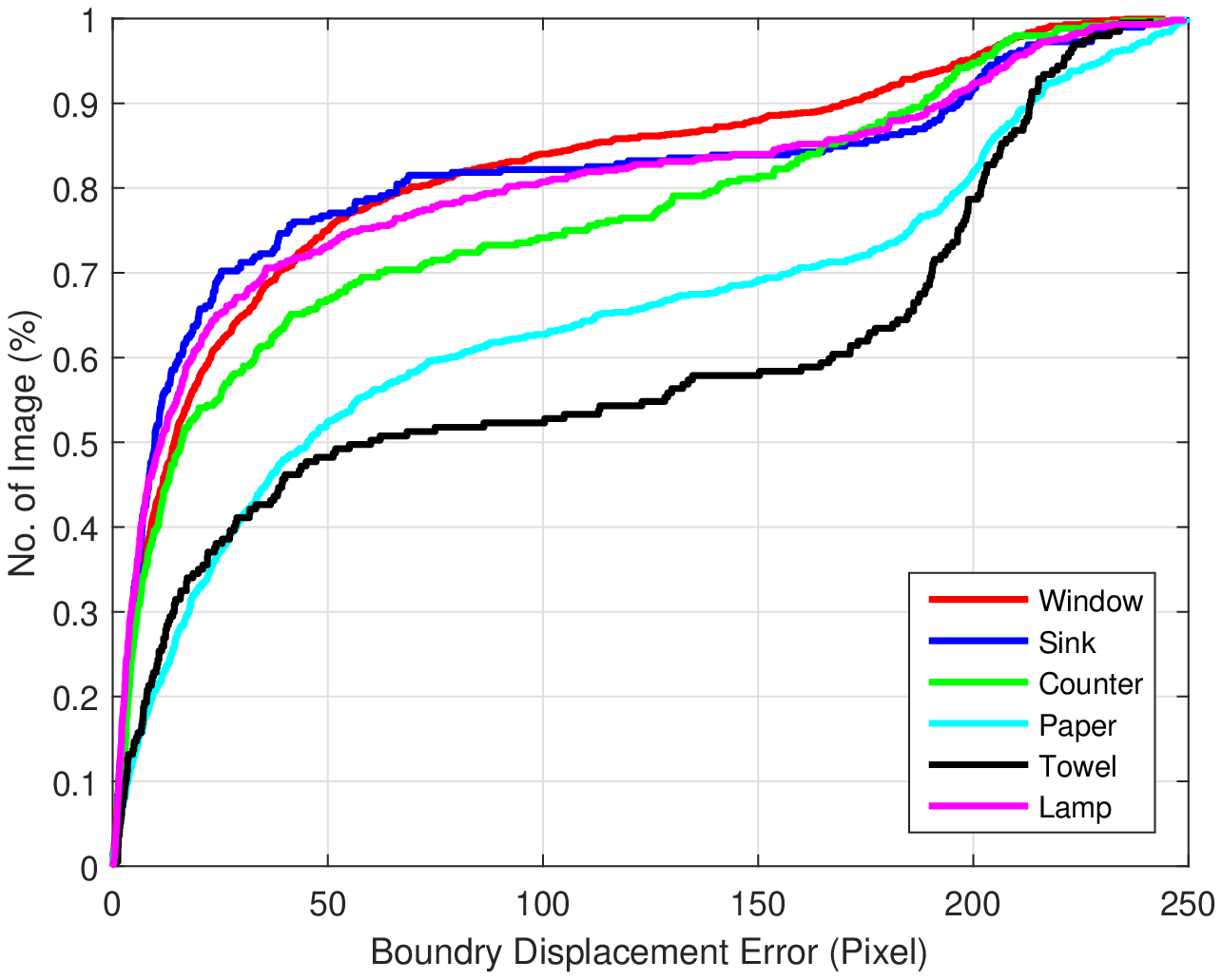}}\vspace{0.1cm}
	\subfloat{\includegraphics[width=0.33\linewidth]{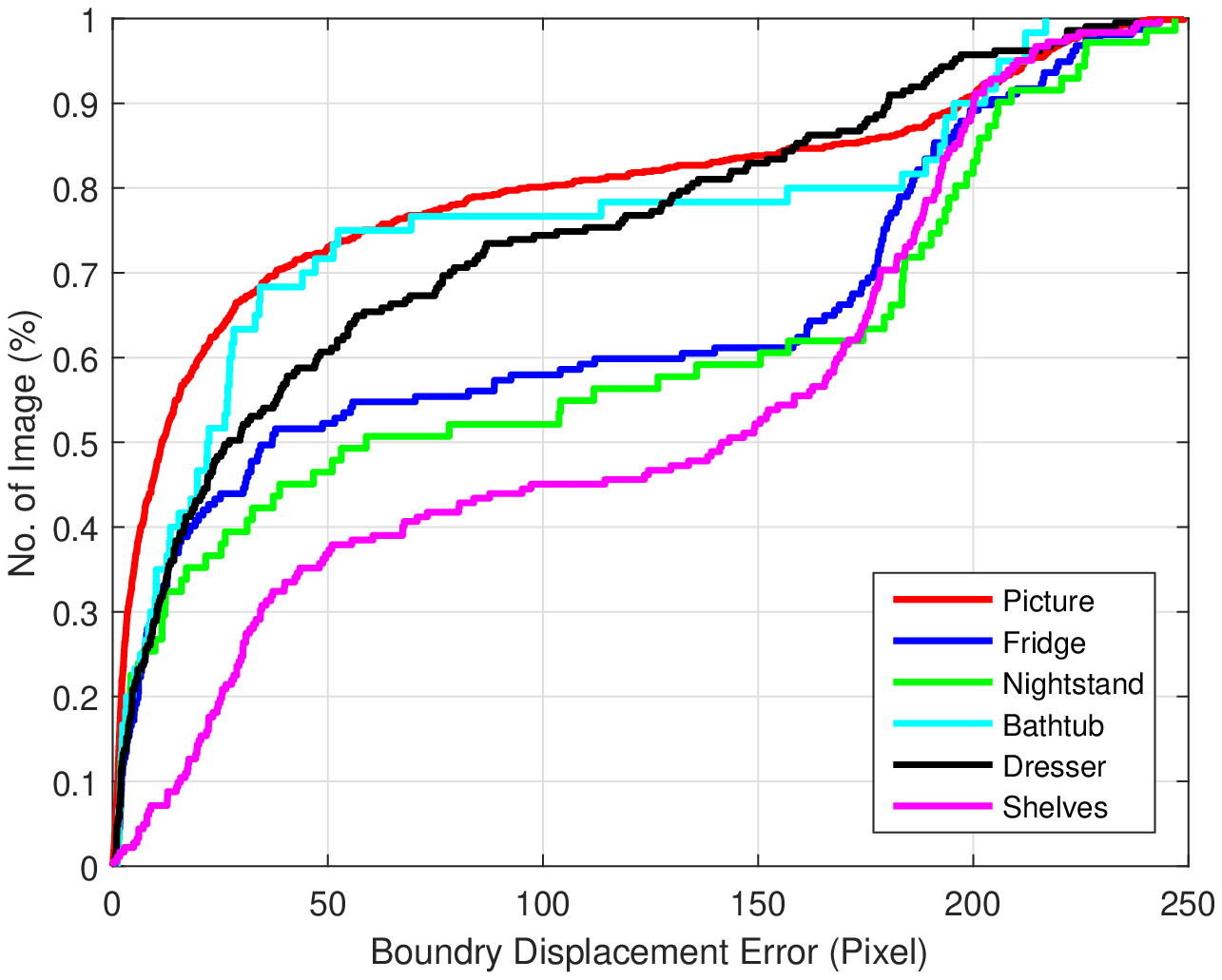}} \hspace{0.05cm}
	\subfloat{\includegraphics[width=0.33\linewidth]{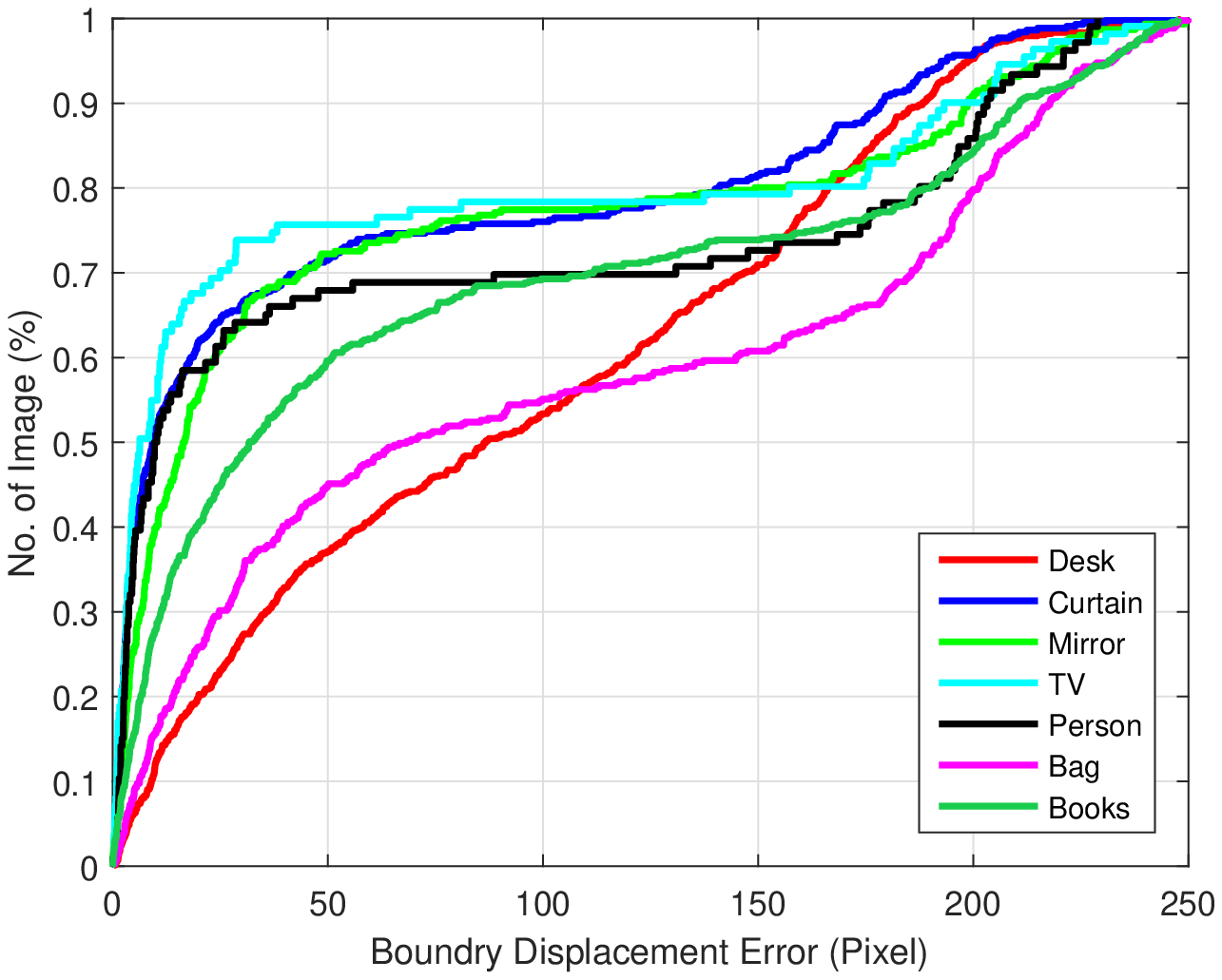}}\hspace{0.05cm}
	\subfloat{\includegraphics[width=0.33\linewidth]{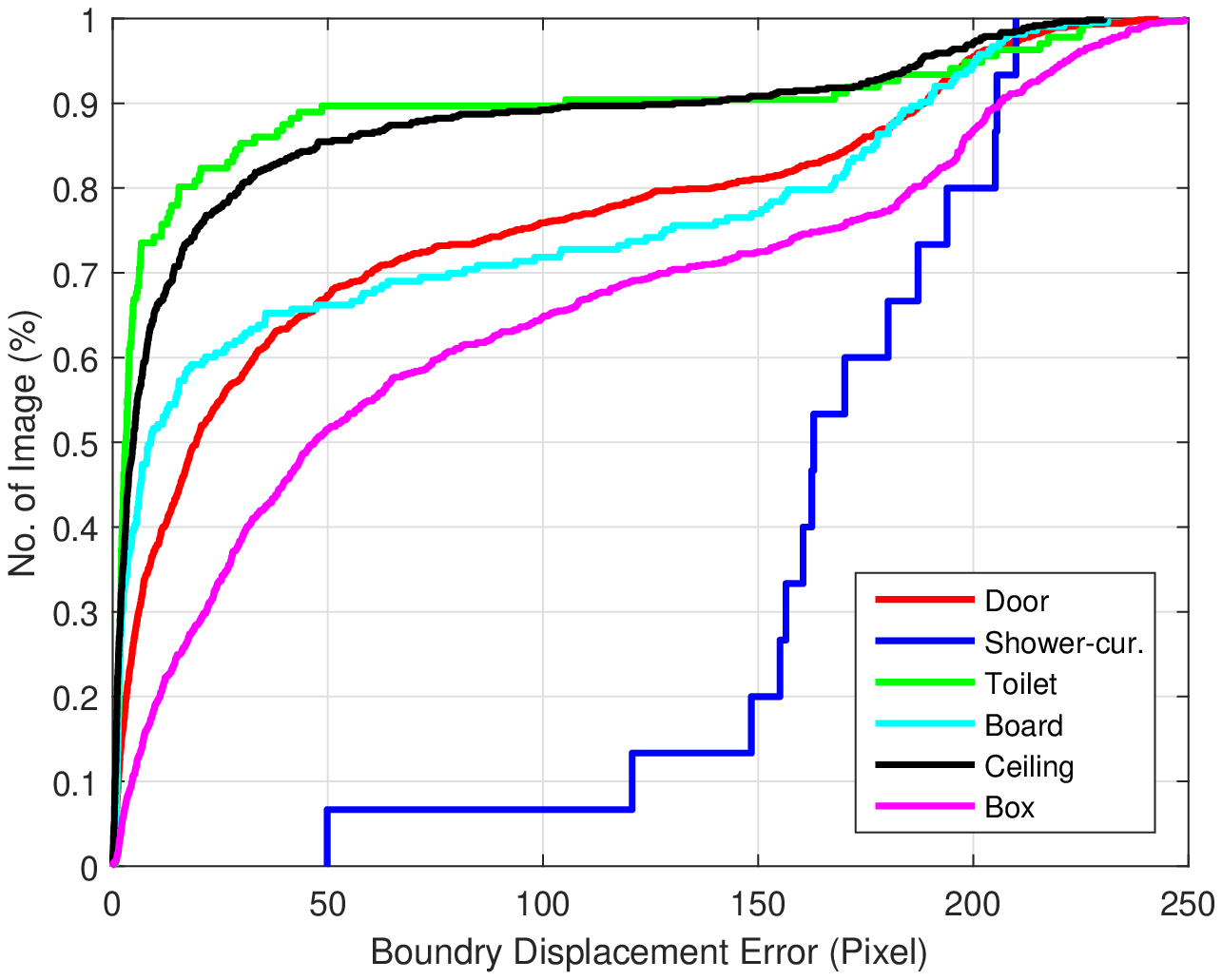}}\vspace{0.1cm}
	\caption{Comparison of boundary displacement error for each class label.}
	\label{fig:bde_per_image}
\end{figure*}

\textbf{Second issue:} Region boundaries are one of the main criteria to determine the quality of image segmentation. The accuracy of these boundaries has not been considered by the global, mean, and IoU metrics, separately. Here, the Boundary Displacement Error (BDE) \cite{freixenet2002yet} was utilized to measure the average displacement error between two segmented boundaries of two images. For each boundary pixel, this error  is defined as the distance between the closest pixel in the other boundary image. This metric has been presented for image segmentation where here we propose to utilize it for each segmented region that belongs to the same class label in the ground-truth and the prediction image. Suppose $B^P$ is the boundary points of a region with class label $l$ in a prediction image and $B^G$ is its corresponding boundary points in the ground-truth image. The two distance distributions are computed from $B^G$ to $B^P$ and from $B^P$ to $B^G$. Then, the minimum distance of each point of $B^P$ from $B^G$ is considered as  $d(x,B^G) = min \{d_E(x,y)\}\ \forall \ y \ in \ B^P$, where $d_E$ is an Euclidean distance. 
To apply this metric in semantic segmentation, the BDE is computed for each class label, separately. Figure \ref{fig:bde_per_image} illustrates the CDF of BDE for each class label. For instance, $\%60$ of images have less than $10$ pixels discrepancy for ‘Floor’ and ‘Chair’ classes (see Figure \ref{fig:bde_per_image}).

\begin{figure*}
	\subfloat[RGB]{\hspace{2.2cm}} \subfloat[Depth] {\hspace{2.2cm}}\subfloat[GT]{\hspace{2.2cm}}\subfloat[Prediction]{\hspace{2.2cm}}\subfloat[RGB]{\hspace{2.4cm}} \subfloat[Depth] {\hspace{2.2cm}}\subfloat[GT]{\hspace{2.4cm}}\subfloat[Prediction]{\hspace{2.2cm}}\vspace{0.1cm}
	\subfloat{\includegraphics[width=0.12\linewidth]{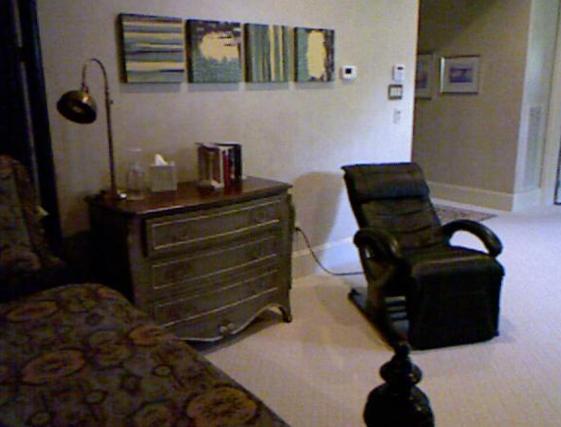}} \hspace{0.001cm}
	\subfloat{\includegraphics[width=0.12\linewidth]{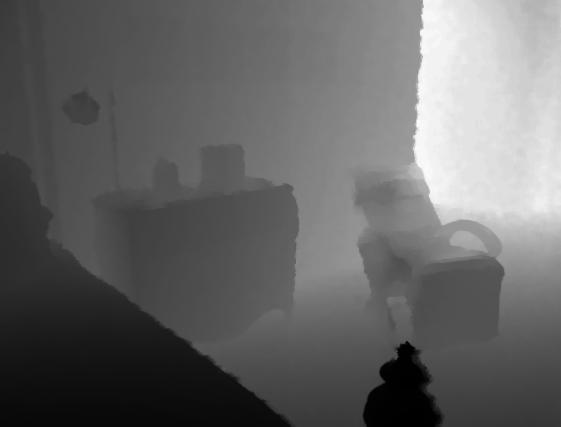}}\hspace{0.001cm}
	\subfloat{\includegraphics[width=0.12\linewidth]{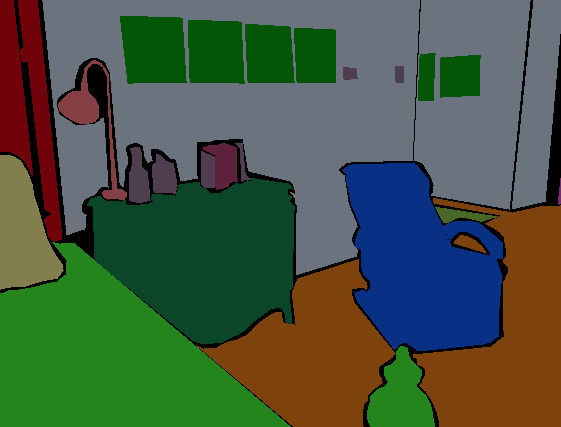}}\hspace{0.001cm}
	\subfloat{\includegraphics[width=0.12\linewidth]{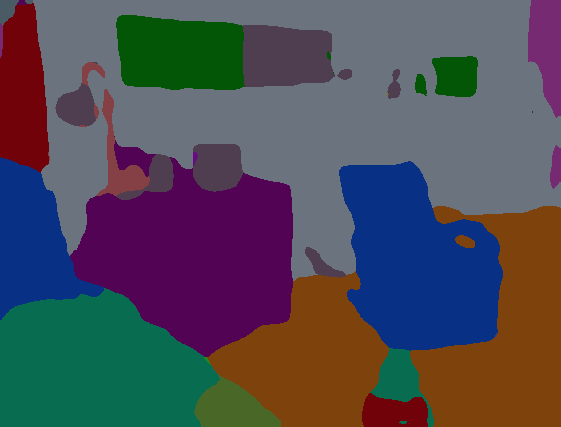}} \hspace{0.01cm}
	\subfloat{\includegraphics[width=0.12\linewidth]{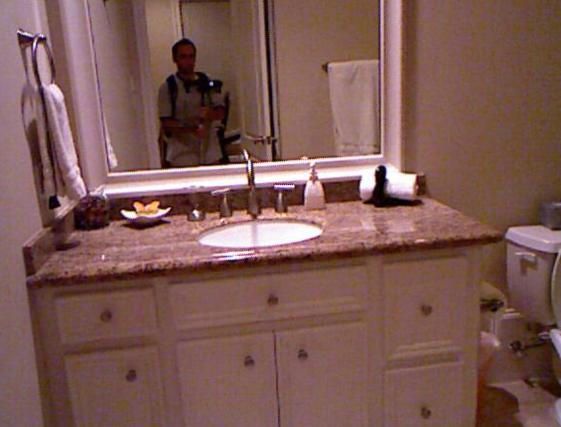}}  \hspace{0.001cm}
	\subfloat{\includegraphics[width=0.12\linewidth]{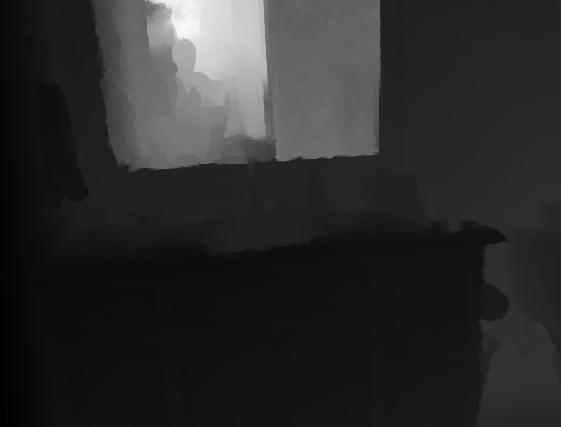}} \hspace{0.001cm}
	\subfloat{\includegraphics[width=0.12\linewidth]{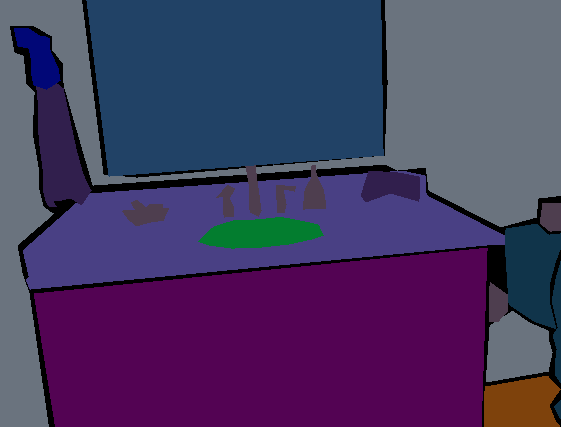}} \hspace{0.001cm}
	\subfloat{\includegraphics[width=0.12\linewidth]{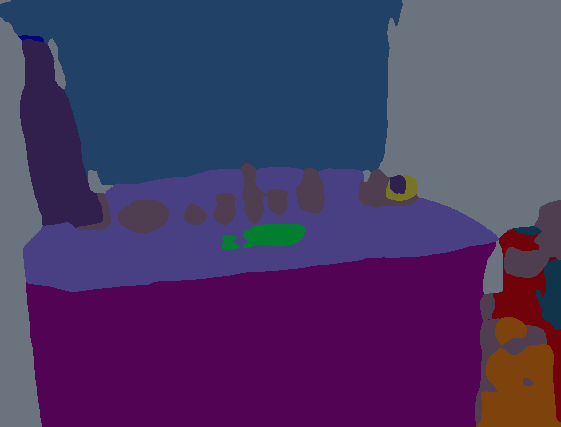}}  \vspace{0.1cm}
	\subfloat{\includegraphics[width=0.12\linewidth]{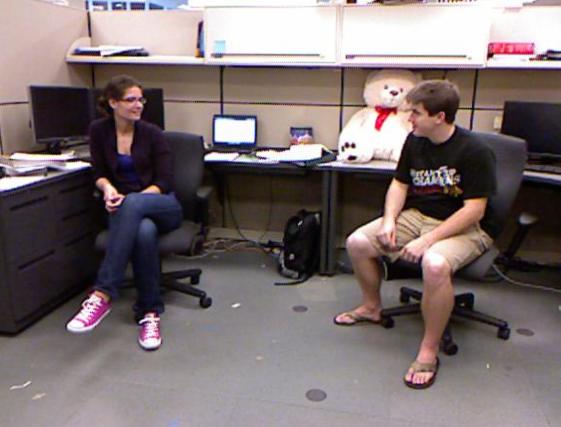}}  \hspace{0.001cm}
	\subfloat{\includegraphics[width=0.12\linewidth]{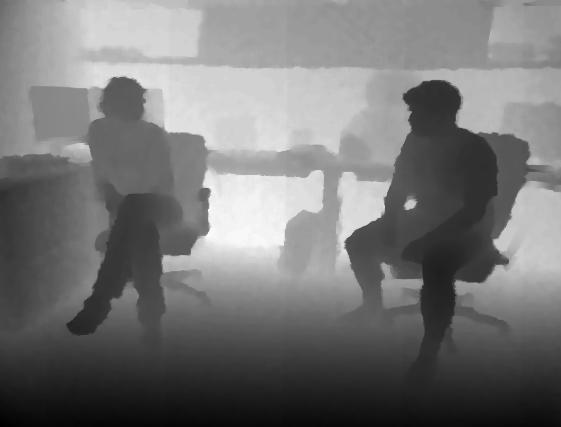}} \hspace{0.001cm}
	\subfloat{\includegraphics[width=0.12\linewidth]{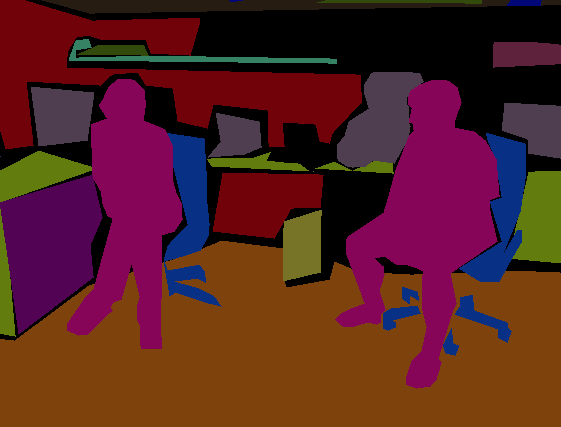}} \hspace{0.001cm}
	\subfloat{\includegraphics[width=0.12\linewidth]{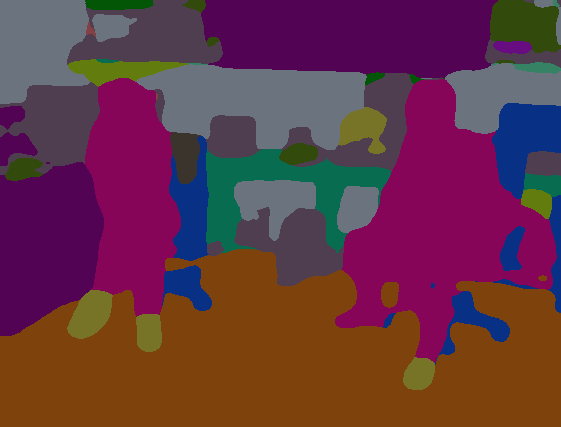}}  \hspace{0.01cm}
	\subfloat{\includegraphics[width=0.12\linewidth]{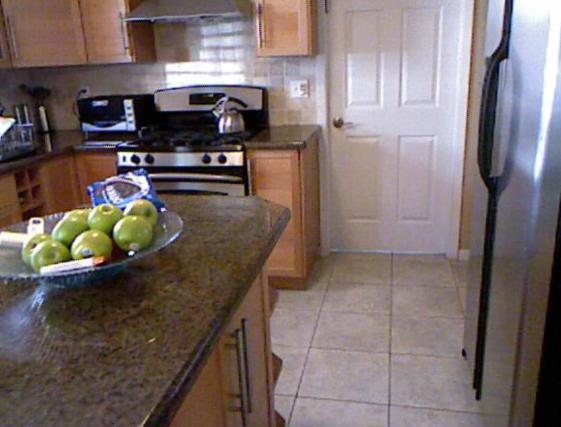}}  \hspace{0.001cm}
	\subfloat{\includegraphics[width=0.12\linewidth]{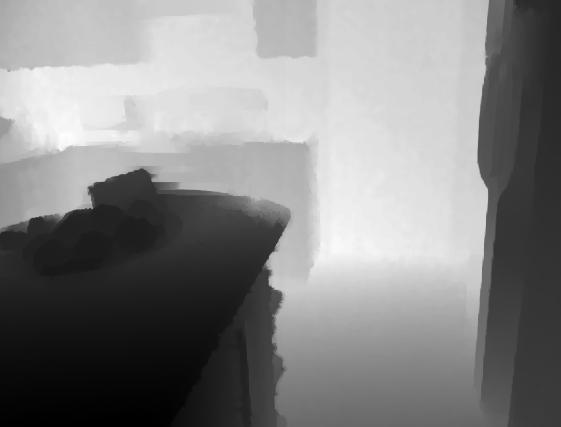}} \hspace{0.001cm}
	\subfloat{\includegraphics[width=0.12\linewidth]{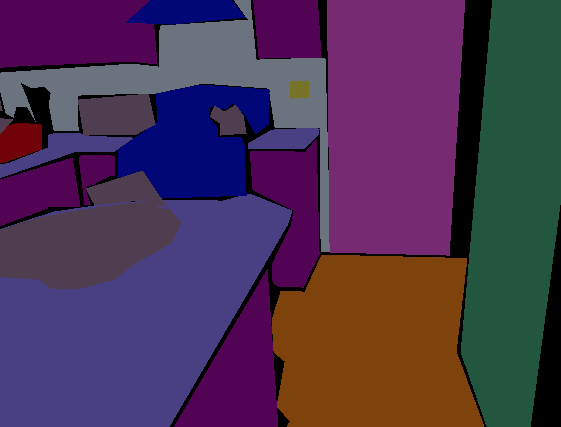}} \hspace{0.001cm}
	\subfloat{\includegraphics[width=0.12\linewidth]{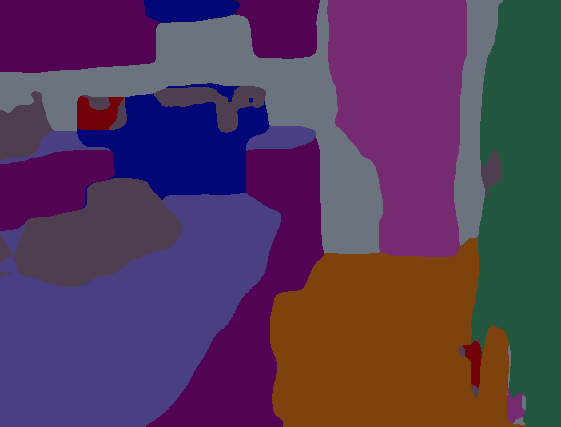}}  \vspace{0.1cm}
	\subfloat{\includegraphics[width=0.12\linewidth]{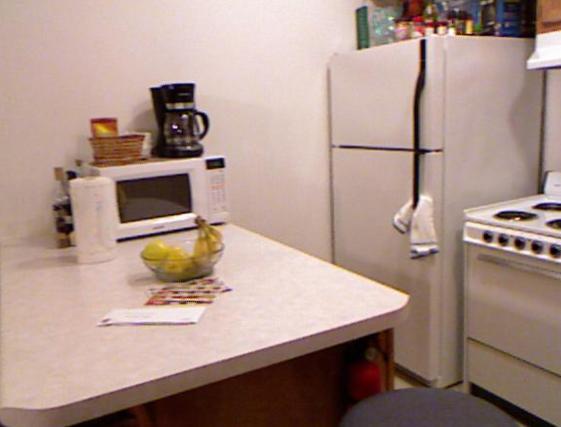}}  \hspace{0.001cm}
	\subfloat{\includegraphics[width=0.12\linewidth]{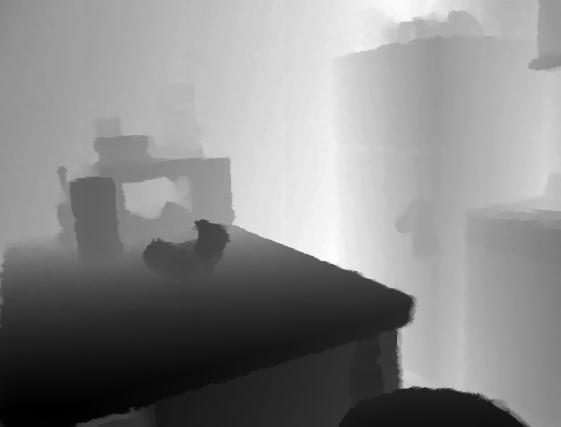}} \hspace{0.001cm}
	\subfloat{\includegraphics[width=0.12\linewidth]{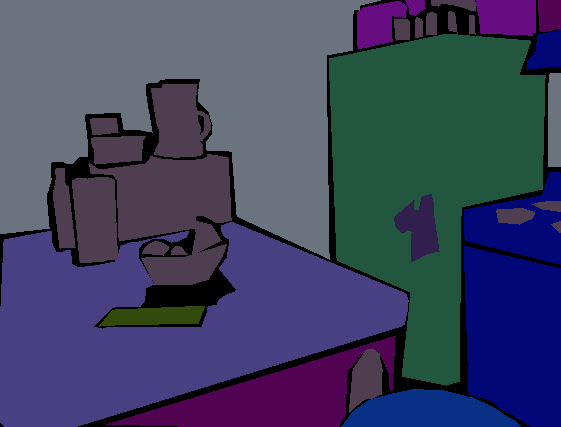}} \hspace{0.001cm}
	\subfloat{\includegraphics[width=0.12\linewidth]{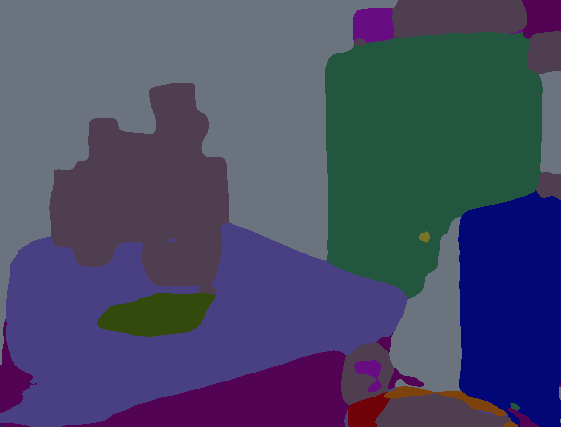}} \hspace{0.01cm}
	\subfloat{\includegraphics[width=0.12\linewidth]{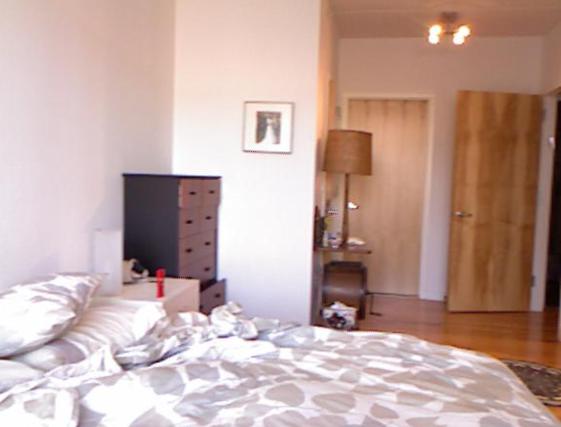}}  \hspace{0.001cm}
	\subfloat{\includegraphics[width=0.12\linewidth]{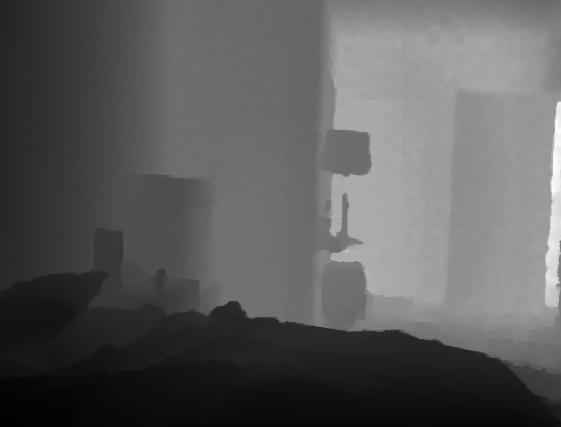}} \hspace{0.05cm}
	\subfloat{\includegraphics[width=0.12\linewidth]{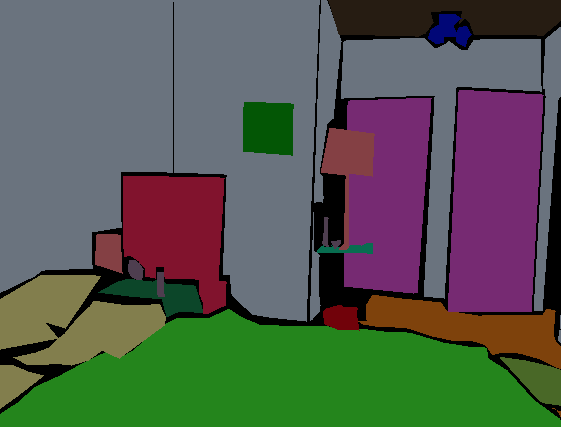}} \hspace{0.001cm}
	\subfloat{\includegraphics[width=0.12\linewidth]{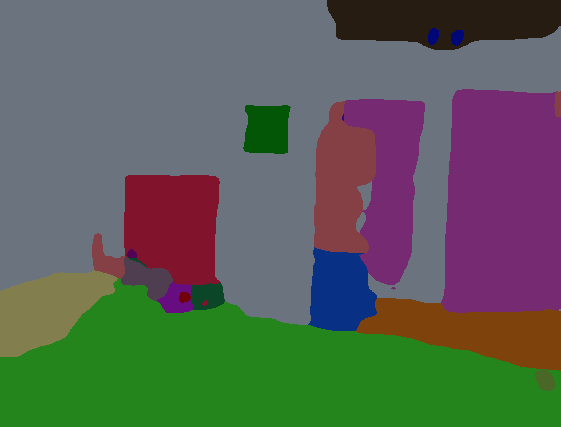}}  \vspace{0.1cm}
	\subfloat{\includegraphics[width=1\linewidth]{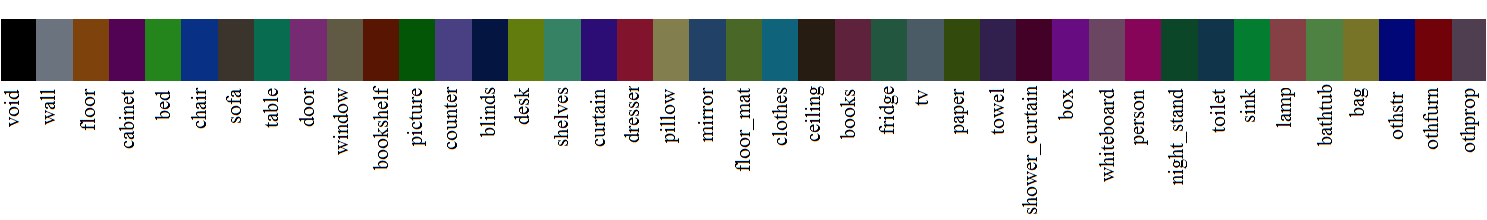}}  \vspace{0.1cm}
	\caption{Qualitative assessments of proposed method on NYU-V2 dataset.}
	\label{fig:nyu_sampel_results}
\end{figure*}
\section{Conclusion}
\label{sec:5}
An efficient attention-based fusion method for RGB and depth fusion was proposed. The proposed  method focused on salient feature maps generated from RGB and depth encoder branches and suppressed unnecessary ones to efficiently fuse these two modalities. The network model was a type of encoder-decoder CNN architectures with two encoder branches and one decoder. The decoder goal was to refine the resolution loss caused by the down sampling procedures in encoder branches via fusion of long-range residual connections coming from both encoder branches. The proposed architecture achieved approximately comparable accuracy  in terms of the IoU score, mean accuracy, and global accuracy, with RGB-D state-of-the-art methods. This same level of accuracy attained  remarkably with  $\%50$ better performance in terms of model size alongside less computational cost (approximately 250G less floating points operations). 
\ifCLASSOPTIONcaptionsoff
  \newpage
\fi



\bibliographystyle{IEEEtran}
\bibliography{attention_fusion}
\end{document}